\documentclass[a4paper,fleqn]{cas-dc}

\usepackage[numbers]{natbib}
\usepackage[table]{xcolor}
\usepackage{multirow}
\usepackage{makecell}
\usepackage{enumitem}

\usepackage{amsmath,amsfonts,bm}

\def\eqref#1{equation~\ref{#1}}

\def\1{\bm{1}}

\def\ve{{\bm{e}}}

\def\vv{{\bm{v}}}
\def\vw{{\bm{w}}}

\DeclareMathAlphabet{\mathsfit}{\encodingdefault}{\sfdefault}{m}{sl}
\SetMathAlphabet{\mathsfit}{bold}{\encodingdefault}{\sfdefault}{bx}{n}

\def\tsc#1{\csdef{#1}{\textsc{\lowercase{#1}}\xspace}}
\tsc{WGM}
\tsc{QE}
\tsc{EP}
\tsc{PMS}
\tsc{BEC}
\tsc{DE}

\begin{document}
\let\WriteBookmarks\relax
\def\floatpagepagefraction{1}
\def\textpagefraction{.001}
\shorttitle{}
\shortauthors{Gao et~al.}

\title [mode = title]{Neuroscience-inspired Staged Representation Learning with Disentangled Coarse- and Fine-Grained Semantics for EEG Visual Decoding}

\author{Xiang Gao}
\credit{Conceptualization, Methodology, Software, Formal analysis, Investigation, Writing -- original draft, Visualization}

\author{Hui Tian}[orcid=0000-0002-7952-571X]
\cortext[cor1]{Corresponding author}
\cormark[1]
\ead{hui.tian@griffith.edu.au}
\credit{Validation, Supervision, Conceptualization}

\author{Yanming Zhu}
\credit{Supervision, Conceptualization, Visualization, Methodology, Writing -- review \& editing}

\author{Xuefei Yin}
\credit{Supervision, Conceptualization, Methodology, Writing -- review \& editing}

\author{Alan Wee-Chung Liew}
\credit{Supervision, Conceptualization, Methodology}

\affiliation{organization={School of Information and Communication Technology, Griffith University},
                addressline={Parklands Drive},
                city={Southport},
                postcode={4222},
                state={QLD},
                country={Australia}}

\begin{abstract}
Decoding visual information from electroencephalography (EEG) signals remains a fundamental challenge in brain-computer interfaces and medical rehabilitation. Existing EEG visual decoding methods mainly focus on learning a single global EEG embedding for cross-modal alignment, but they largely overlook the staged and hierarchical characteristics of human visual processing. To address this limitation, we propose a neuroscience-inspired staged representation learning framework that reformulates EEG visual decoding as a stage-specific representation decomposition problem. The proposed framework organizes EEG representation learning into three complementary phases, including low-level visual representation learning, high-level semantic representation learning, and integrative information fusion. To strengthen semantic modeling, we further introduce a multimodal dual-level semantic learning mechanism that separates coarse label-level semantics from fine image-level visual-semantic information. In addition, semantic latent channels are introduced as computational representation channels generated from observed visual EEG signals, expanding the channel-level semantic representation space for structured semantic abstraction and cross-modal alignment. 
Extensive experiments on the THINGS-EEG benchmark demonstrate that the proposed method achieves superior performance under subject-dependent zero-shot evaluation and improved exact retrieval under subject-independent zero-shot evaluation. Additional analyses, including layer-wise retrieval, temporal accumulation, expanded multi-image retrieval, and ablation studies, further support the effectiveness of the staged decomposition and structured semantic modeling strategy. These results suggest that explicitly modeling staged perceptual, semantic, and integrative representations provides an effective neuroscience-inspired framework for EEG-based visual decoding.
\end{abstract}


\maketitle

\section{Introduction}
Decoding visual information from electroencephalography (EEG) signals is a central task in brain-computer interface research \citep{wilson2024feasibility, ferrante2024decoding_vs}, with broad implications for neurorehabilitation \citep{zhang2025cat}, visual cognition \citep{de2024perceptual}, and brain-inspired artificial intelligence \citep{pereira2018toward, ding2025eeg}. By learning mappings between neural activity and visual content, EEG visual decoding provides a useful framework for understanding how the human brain represents perceptual and semantic information, while also supporting the development of practical systems that infer visual intent or cognitive states from brain signals. Most existing EEG visual decoding methods therefore aim to learn informative neural representations that can be aligned with visual or semantic features, thereby enabling the recovery of visual information from brain activity.

Recent progress in EEG-based visual decoding has been largely driven by cross-modal alignment. Existing methods mainly improve EEG embeddings for alignment with visual or multimodal semantic features through noise-aware priors, multimodal feature enrichment, semantic consistency regularization, graph-based modeling, or vision-language supervision \citep{wu2025bridging, zhang2025cognitioncapturer, zhang2025category, li2024visual, chen2024visual, du2023decoding, ferrante2024decoding, song2025recognizing}. These studies have improved the quality of EEG representations and advanced EEG-based visual retrieval and recognition. Nevertheless, despite their differences in encoder design, auxiliary supervision, and optimization objectives, many existing methods still share a similar end goal. They mainly seek to learn a more discriminative final EEG representation for cross-modal matching, while richer priors or supervision are typically incorporated within a largely unified embedding framework rather than being organized into explicitly structured stages.

However, human visual perception is not a single-stage process. Neuroscientific evidence suggests that visual information is processed progressively and hierarchically, evolving from early perceptual analysis to higher-level semantic abstraction and subsequent information integration \citep{felleman1991distributed, goodale1992separate, graumann2022spatiotemporal, kappenman2021erp, xu2021review}. This perspective reveals two important limitations that remain insufficiently addressed in existing EEG visual decoding methods. First, low-level perceptual cues, high-level semantic information, and final decision-oriented representations are often compressed into a single shared embedding, rather than being modeled in an explicitly staged manner. As a result, the progressive transformation from perceptual processing to semantic understanding is not directly reflected in the learned EEG representations. Second, although semantic supervision has been increasingly introduced, it is still often incorporated without explicitly distinguishing coarse label-level semantics from fine image-level semantics. This may obscure the complementary roles of category-level abstraction and instance-level discrimination, both of which are important for robust EEG visual decoding.

In addition, prior EEG visual decoding methods rarely explore whether semantic modeling can benefit from explicitly structured channel-level representations. This issue is relevant because neuroscientific evidence suggests that EEG channel involvement is not functionally uniform, and that channel specialization may provide useful inductive guidance for representation learning \citep{kroczek2019contributions}. Although such evidence does not directly transfer from language comprehension to visual decoding, it motivates the broader idea that semantic information in EEG may be better modeled with an explicitly designed channel-level representation space, rather than relying solely on the observed channels and a single undifferentiated global embedding.

Motivated by these observations, we propose a staged representation learning framework for EEG visual decoding, guided by neuroscientific evidence. Instead of treating EEG visual decoding as a single global embedding alignment problem, the proposed framework reformulates it as a stage-specific representation decomposition problem. Specifically, EEG representation learning is organized into three complementary phases, including low-level visual representation learning, high-level semantic representation learning, and integrative information fusion. This design allows perceptual cues, fine image-level visual-semantic information, and coarse label-level semantics to be modeled in distinct but coordinated representation spaces.
To further strengthen semantic modeling, we introduce a multimodal dual-level semantic learning mechanism that separates coarse label-level semantics from fine image-level semantics. We also introduce semantic latent channels as a learnable channel-structured latent basis for semantic abstraction, rather than as additional physiological electrodes. The whole framework is jointly optimized through a multi-level objective, enabling stage-specific supervision before final cross-modal alignment. Extensive experiments on the THINGS-EEG benchmark \citep{gifford2022large} demonstrate the effectiveness of the proposed framework under both subject-dependent and subject-independent zero-shot evaluation settings. Additional analyses, including layer-wise retrieval, temporal accumulation, expanded multi-image retrieval, and ablation studies, further support the effectiveness of the staged decomposition and structured semantic modeling strategy.

The main contributions are summarized as follows:
\begin{itemize}[itemsep=0pt, topsep=0pt]
    \item We reformulate EEG visual decoding as a stage-specific representation decomposition problem, where low-level perceptual representation learning, high-level semantic representation learning, and integrative information fusion are jointly optimized within a unified cross-modal learning framework.

    \item We design a multimodal dual-level semantic learning mechanism that separates coarse label-level semantics from fine image-level visual-semantic information. The coarse branch is supervised by text features derived from image labels, while the fine branch is supervised by high-level image features, enabling structured semantic alignment beyond conventional global embedding learning.

    \item We introduce semantic latent channels as computational representation channels generated from observed visual EEG signals. These latent channels expand the channel-level semantic representation space and provide a structured basis for coarse semantic abstraction and fine image-level semantic interaction.

	\item We conduct extensive experiments on the THINGS-EEG benchmark under both subject-dependent and subject-independent zero-shot settings, demonstrating the effectiveness of the proposed framework for EEG visual decoding. Additional analyses, including layer-wise retrieval, temporal accumulation, expanded multi-image retrieval, and ablation studies, further support the effectiveness of the staged design and the proposed semantic modeling strategy.
\end{itemize}

\section{Related Work}
\label{sec:related_work}

\subsection{EEG Decoding via Cross-Modal Alignment}

Recent progress in EEG-based visual decoding has been largely driven by cross-modal alignment, where EEG representations are learned to match visual embeddings for retrieval or recognition. Early large-scale evidence was provided by Du et al.~\citep{du2023decoding}, who jointly modeled brain, visual, and linguistic features to bridge EEG signals and semantic representations. Song et al.~\citep{song2024decoding} further showed that self-supervised contrastive learning can learn transferable EEG representations for natural image recognition. These studies established cross-modal representation learning as a dominant paradigm for EEG visual decoding.
Building on this paradigm, many recent methods have focused on improving the quality of EEG representations or the effectiveness of the alignment objective. Wu et al.~\citep{wu2025bridging} introduced an uncertainty-aware blur prior to alleviate the effect of noisy neural responses. Zhang et al.~\citep{zhang2025cognitioncapturer} incorporated multimodal priors such as texture and depth to enrich visual alignment, while Zhang et al.~\citep{zhang2025category} leveraged wavelet-domain modeling and semantic-aware contrastive supervision to improve category discrimination. Chen et al.~\citep{chen2024visual} strengthened visual-EEG matching through semantic consistency regularization, and Shi et al.~\citep{shi2025brainalign} modeled frequency-aware temporal structure and differentiable cluster assignment for more discriminative EEG embeddings. Although these approaches differ in architectural design and supervision strategy, they largely remain within a cross-modal alignment paradigm that emphasizes learning a stronger final EEG representation for matching.

\subsection{Semantic and Generative Extensions}
Beyond improving alignment architectures directly, a related line of work enhances EEG visual decoding with richer supervision signals or auxiliary multimodal priors. Ferrante et al.~\citep{ferrante2024decoding} distilled knowledge from CLIP to transfer vision-language semantics into EEG representations. Song et al.~\citep{song2025recognizing} further introduced language-guided contrastive learning, highlighting the value of textual semantics for natural image recognition from EEG. Rajabi et al.~\citep{rajabi2025human} showed that human-aligned image priors can improve visual decoding from brain signals, and Ma and Ruotsalo~\citep{ma2024cognition} explored cognition-supervised alignment between EEG responses and visual saliency. These studies suggest that richer supervision beyond raw visual features can improve the semantic quality of learned EEG representations.
In parallel, generative approaches have extended EEG visual decoding from recognition-oriented retrieval to image reconstruction. Zeng et al.~\citep{zeng2023dm} proposed a diffusion-based framework for EEG-to-image reconstruction. Li et al.~\citep{li2024visual} combined EEG embeddings with guided diffusion for both decoding and reconstruction, and Xiao et al.~\citep{xiao2025eeg} modeled 3D geometric structure together with nonstationarity for EEG decoding and visual reconstruction. Although these methods broaden the task scope and enrich the forms of supervision, they still largely center on a single final EEG latent representation or a dominant end-stage alignment objective, rather than explicitly modeling how perceptual and semantic representations evolve across stages.

\subsection{Neuroscience-Inspired Modeling and Gaps}
\label{subsec:neuroscience_motivation}

Despite their differences in encoder design, auxiliary priors, and training objectives, most existing methods still center on learning a stronger final EEG representation for cross-modal matching. Low-level perceptual cues, image-level semantics, and label-level conceptual information are usually compressed into one shared embedding, or are only coupled indirectly through auxiliary losses. As a result, the progressive nature of human visual processing is rarely modeled explicitly in EEG visual decoding.
This limitation is important because neuroscientific studies suggest that visual perception unfolds progressively and hierarchically. Felleman and Van Essen~\citep{felleman1991distributed} described the distributed hierarchical organization of the primate visual cortex, while Goodale and Milner~\citep{goodale1992separate} proposed the influential two-visual-pathway hypothesis. Recent EEG and neuroimaging studies further show that object-related visual information is represented dynamically over time~\citep{graumann2022spatiotemporal}, and ERP research indicates that EEG responses reflect different cognitive processes across temporal stages~\citep{kappenman2021erp,xu2021review}. These findings motivate a representation learning formulation in which perceptual, semantic, and integrative information are modeled as related but distinct components, rather than being collapsed into a single final embedding.

Another limitation is that existing EEG visual decoding methods rarely consider whether semantic modeling can benefit from explicitly structured channel-level representations. Neuroscientific evidence suggests that EEG channel involvement is not functionally uniform, and that frontal and temporal regions can contribute differently to semantic processing in language-comprehension tasks~\citep{kroczek2019contributions}. However, such evidence should not be interpreted as a direct physiological mapping for visual decoding. Instead, it motivates a more general computational hypothesis: semantic information in EEG may be more effectively modeled when the representation space provides a structured channel-level basis for semantic abstraction, rather than relying only on the observed visual channels and a single undifferentiated global embedding.

Motivated by these gaps, we propose a neuroscience-inspired staged representation learning framework for EEG visual decoding. The framework is not intended to reproduce the biological visual system directly. Rather, it uses neuroscientific evidence as design motivation for organizing EEG representation learning into low-level visual learning, high-level semantic learning, and integrative fusion. Within this formulation, dual-level semantic modeling and semantic latent channels are introduced to support structured alignment with both image-level and label-level semantics.

\section{Proposed Method}
\label{sec:method}
\begin{figure*}[t]
\begin{center}
\includegraphics[width=\textwidth]{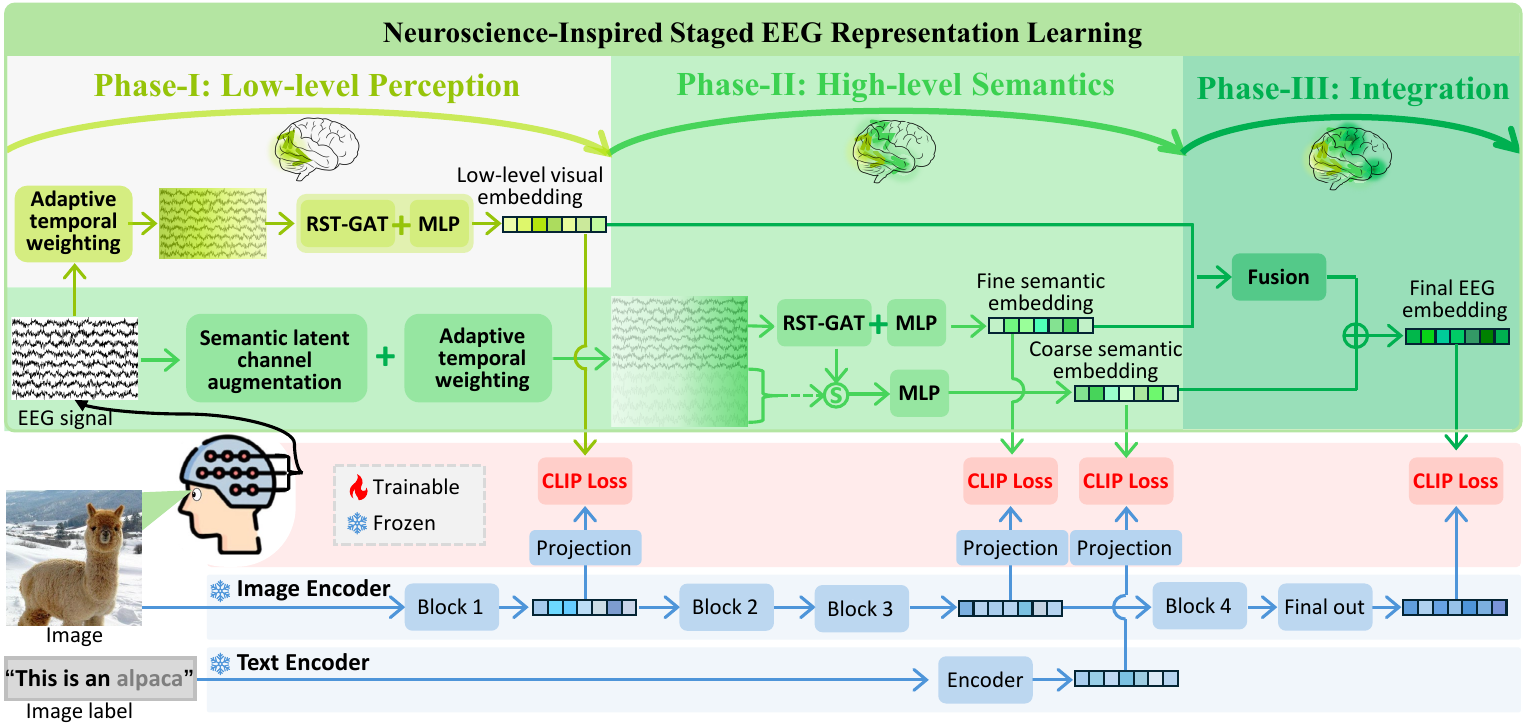} 
\end{center}
\caption{Overview of the proposed neuroscience-inspired staged EEG representation learning framework. \textbf{Phase-I} learns low-level visual representations from 17 visual-related EEG channels and aligns them with low-level image features. \textbf{Phase-II} introduces semantic latent channels and performs dual-level semantic learning, where coarse semantics are aligned with text features derived from image labels and fine image-level visual-semantic representations are aligned with high-level image features. \textbf{Phase-III} integrates the low-level visual representation and the structured semantic representations into a unified EEG embedding for final image alignment. The entire framework is optimized with multi-level CLIP losses to enforce cross-modal consistency across stages.}
\label{fig:framework}
\end{figure*}

The overall architecture of the proposed framework is illustrated in Fig.~\ref{fig:framework}. Motivated by the neuroscience-inspired design considerations discussed in Section~\ref{subsec:neuroscience_motivation}, our method models EEG visual decoding as a staged representation learning problem. The goal is not to impose a strict biological correspondence between model components and neural processing stages, but to use staged organization as an inductive bias for decomposing EEG representations into perceptual, semantic, and integrative components. Specifically, the framework progressively transforms neural signals from low-level perceptual representations to higher-level semantic representations and finally to an integrated EEG representation for cross-modal alignment. Here, the staged design refers to a structured decomposition of representation learning and supervision, rather than sequential phase-by-phase optimization. The framework consists of three phases that are jointly optimized end-to-end through the multi-level objective in Eq.~\ref{eq:loss_total}: \textbf{Phase-I} learns low-level visual representations from visual-related EEG channels and aligns them with low-level image features; \textbf{Phase-II} models high-level semantics through a multimodal dual-level semantic learning mechanism that disentangles coarse label-level semantics and fine image-level semantics; and \textbf{Phase-III} integrates the learned perceptual and semantic representations into a unified EEG embedding for final image alignment. In addition, to enrich semantic modeling in Phase-II, we introduce semantic latent channels, which are computational representation channels generated from observed visual EEG signals rather than additional physiological electrodes. These semantic latent channels expand the channel-level semantic representation space and support more structured cross-modal alignment. The details of the three phases are described in the following subsections.

\subsection{Phase-I: Low-Level Perception} 
\label{sec:Phase-I}
In Phase-I, our goal is to learn low-level visual representations from EEG signals and align them with low-level image features. The low-level image features are obtained from the \texttt{block1} output of the frozen OpenCLIP ResNet50 image encoder. For EEG, we selectively use 17 channels that are primarily associated with visual responses\footnote{17 channels primarily associated with visual responses: P7, P5, P3, P1, Pz, P2, P4, P6, P8, PO7, PO3, POz, PO4, PO8, O1, Oz, O2~\citep{gifford2022large}.}~\citep{kappenman2021erp, gifford2022large}. Motivated by evidence that low-level visual responses are concentrated in relatively early temporal windows~\citep{goodale1992separate, graumann2022spatiotemporal}, we introduce a learnable adaptive temporal weighting mechanism that allows the model to emphasize temporally informative EEG segments during representation learning.

Formally, let $(\mathbf{e}_v, \mathbf{v})$ denote a batch of EEG-image pairs, where $\mathbf{e}_v \in \mathbb{R}^{B \times C_1 \times T}$ represents $B$ EEG samples with $C_1 = 17$ channels and $T$ time steps, and $\mathbf{v} \in \mathcal{V}$ denotes the corresponding images. To emphasize temporally informative EEG segments, we use a learnable temporal weighting module to produce a normalized temporal weight vector $\mathbf{w}_t \in \mathbb{R}^{1 \times T}$. As summarized in Table~\ref{tab:full_backbone_setting}, this module is implemented as Linear$(T \rightarrow 128)$ + ReLU + Linear$(128 \rightarrow T)$ + Softmax. The resulting temporal weights are shared across EEG channels and broadcast along the channel dimension during element-wise reweighting.

For low-level EEG visual representation learning, we design a \textbf{Residual SpatioTemporal Graph Attention Network (RST-GAT)} to encode the low-level EEG visual representation $\mathbf{e}_1 \in \mathbb{R}^{B \times 256}$. The term \emph{spatiotemporal} here refers to graph-based modeling over both EEG channels and temporal indices. Given an EEG representation $\mathbf{x}\in\mathbb{R}^{B\times C\times T}$, RST-GAT models it from two complementary graph views. In the channel-wise graph view, each EEG channel is treated as a graph node, and its $T$-dimensional temporal response is used as the node feature. This allows the channel graph to capture dependencies among EEG channels. In the temporal-wise graph view, each temporal index is treated as a graph node, and its $C$-dimensional channel response is used as the node feature. This allows the temporal graph to capture dependencies among time steps. The outputs of the graph attention operations are incorporated through residual updates and then mapped to the embedding space through an MLP, as summarized in Table~\ref{tab:full_backbone_setting}. In Phase-I, $C=17$, while the same RST-GAT formulation is later applied to the expanded Phase-II representation with $C=29$. This encoding process is defined as
\begin{equation}
\label{eq:phase-I}
\ve_1
=
f_{\mathrm{MLP}}\Big(
f_{\mathrm{ST\text{-}GAT}}(\ve_v \otimes \vw_t + \ve_v) + \ve_v
\Big),
\end{equation}
where $\otimes$ denotes element-wise multiplication with broadcasting along the channel dimension. 

Meanwhile, the low-level image feature $\mathbf{v}_1 \in \mathbb{R}^{B \times 256}$ is obtained by projecting the \texttt{block1} output of the frozen OpenCLIP ResNet50 image encoder through an MLP. To encourage alignment between the low-level EEG embeddings and the corresponding low-level image features, the Phase-I alignment loss is defined as
\begin{equation}
\mathcal{L}^{I}_{\mathrm{CLIP}} = \mathrm{CLIP\_LOSS}(\ve_1, \vv_1).
\label{eq:clip_l}
\end{equation}
Here, $\mathrm{CLIP\_LOSS}(\cdot,\cdot)$ denotes a symmetric CLIP-style contrastive loss. Given two batches of paired embeddings $\mathbf{X}=[\mathbf{x}_1,\ldots,\mathbf{x}_B]^\top \in \mathbb{R}^{B\times d}$ and $\mathbf{Y}=[\mathbf{y}_1,\ldots,\mathbf{y}_B]^\top \in \mathbb{R}^{B\times d}$, we first apply $L_2$ normalization to obtain $\bar{\mathbf{x}}_i$ and $\bar{\mathbf{y}}_i$. The pairwise similarity is computed as
\begin{equation}
\label{eq:clip_similarity}
s_{ij}
=
\lambda \bar{\mathbf{x}}_i^\top \bar{\mathbf{y}}_j,
\end{equation}
where $\lambda=\mathrm{Softplus}(\rho)$ is a learnable positive logit scale. The symmetric CLIP-style contrastive loss is then formulated as
\vspace{-0.2in}
\begin{equation}
\label{eq:clip_loss}
\mathrm{CLIP\_LOSS}(\mathbf{X},\mathbf{Y})
=
\frac{1}{2}
\left(
\mathcal{L}_{X\rightarrow Y}
+
\mathcal{L}_{Y\rightarrow X}
\right),
\end{equation}
where
\begin{equation}
\label{eq:clip_loss_direction}
\left\{
\begin{aligned}
\mathcal{L}_{X\rightarrow Y}
&=
-\frac{1}{B}
\sum_{i=1}^{B}
\log
\frac{\exp(s_{ii})}
{\sum_{j=1}^{B}\exp(s_{ij})},
\\
\mathcal{L}_{Y\rightarrow X}
&=
-\frac{1}{B}
\sum_{i=1}^{B}
\log
\frac{\exp(s_{ii})}
{\sum_{j=1}^{B}\exp(s_{ji})}.
\end{aligned}
\right.
\end{equation}

\subsection{Phase-II: High-Level Semantics}
\label{sec:Phase-II}

Within the proposed staged framework, Phase-II is designed to model the stage at which EEG responses increasingly reflect higher-level semantic information. While Phase-I focuses on low-level visual representation learning and Phase-III performs integrative information fusion, Phase-II is responsible for learning structured semantic representations from EEG signals. To this end, we organize semantic representation learning into two complementary levels. \textbf{Coarse semantics} capture abstract category-level distinctions and are aligned with text features derived from class labels, whereas fine image-level semantics capture more detailed visual-semantic information and are aligned with high-level image features. This dual-level design allows the model to account for both category-level semantic abstraction and instance-level visual-semantic specificity within a unified framework.

To support such structured semantic modeling, we further introduce semantic latent channels. These channels are not intended to represent additional physiological electrodes. Instead, they define an explicitly introduced latent channel space in Phase-II, which complements the observed visual EEG channels and provides a more suitable representation basis for semantic abstraction. In particular, the semantic latent channels are used to support coarse semantic modeling, while the joint representation of visual and latent channels is used to support fine semantic modeling.

\subsubsection{Semantic Latent Channel Augmentation}
\label{subsubsec:semantic_latent_channels}

Neuroscientific studies suggest that EEG channel involvement is not functionally uniform, and that a subset of channels can be selectively associated with semantic processing during language comprehension\footnote{Semantic-related channels for language comprehension: Fp1, F3, F7, FC5, FC1, C3, T7, CP5, P7, FT7, F5, TP7~\citep{kroczek2019contributions}.}~\citep{kroczek2019contributions}. However, these findings were obtained in a language-comprehension setting and do not imply that the same channels can be directly treated as semantic channels for visual decoding. In contrast, visual evoked responses are typically recorded most reliably over posterior and occipital scalp regions associated with the visual pathway~\citep{creel2019visually,thigpen2017assessing}.

Motivated by this distinction, we do not directly reuse language-related channels in the visual domain. Instead, we introduce \emph{semantic latent channels} as a learnable latent channel space for semantic abstraction in Phase-II. The choice of 12 latent channels is inspired by prior evidence on semantic-related channel specialization, but it should be understood as a design choice for structuring semantic representation learning rather than as a claim of one-to-one physiological correspondence.

Importantly, these semantic latent channels are not introduced as additional physiological electrodes or unconstrained free channels. Rather, they are initialized from the observed visual EEG through a learnable channel projection applied independently at each time step. Given the visual EEG signal $\mathbf{e}_v \in \mathbb{R}^{B \times 17 \times T}$, the semantic latent channels are generated as
\begin{equation}
\label{eq:SLC_init}
\mathbf{e}_{\mathrm{slc}}(:,:,t)=\mathbf{e}_v(:,:,t)\mathbf{W}_{\mathrm{slc}}^{\top}
+
\mathbf{1}_{B}\mathbf{b}_{\mathrm{slc}}^{\top},
\quad
t = 1, \ldots, T,
\end{equation}
where $\mathbf{W}_{\mathrm{slc}} \in \mathbb{R}^{12 \times 17}$, $\mathbf{b}_{\mathrm{slc}} \in \mathbb{R}^{12}$, and $\mathbf{1}_{B}\in\mathbb{R}^{B\times 1}$ denotes an all-one vector used to broadcast the bias across the batch dimension. In this way, $\mathbf{e}_{\mathrm{slc}} \in \mathbb{R}^{B \times 12 \times T}$ remains grounded in the observed visual EEG while expanding the representation space available for semantic modeling. In the proposed framework, the semantic latent channels generated by Eq.~\ref{eq:SLC_init} mainly support coarse semantic abstraction, while their interaction with the observed visual channels also contributes to finer image-level semantic learning in Section~\ref{subsubsec:dual-level}.

\subsubsection{Dual-Level Semantic Learning Mechanism}
\label{subsubsec:dual-level}

In Phase-II, our goal is to learn structured high-level semantic representations from EEG signals. To reflect the hierarchical nature of semantic processing, we design a multimodal dual-level semantic learning mechanism that organizes EEG semantics into two complementary branches. 
The \textbf{coarse semantic branch} focuses on abstract category-level semantics and is aligned with text features derived from image labels. The \textbf{fine semantic branch} focuses on more detailed image-level visual-semantic information and is aligned with high-level image features.
In this way, Phase-II models semantic abstraction and semantic specificity in a coordinated manner, rather than relying on a single undifferentiated semantic representation.

Motivated by evidence that semantic processing becomes more prominent in relatively later temporal stages of EEG responses~\citep{kappenman2021erp, felleman1991distributed, xu2021review}, we introduce a separate learnable temporal weighting module for Phase-II to adaptively highlight temporally informative signals for semantic representation learning. This module has the same architecture as the Phase-I temporal weighting module summarized in Table~\ref{tab:full_backbone_setting}, but uses independent parameters, allowing semantic representation learning to emphasize temporal evidence differently from low-level visual representation learning. After semantic latent channel initialization, the EEG signals are expanded into a joint channel representation,
\[
\ddot{\mathbf{e}} = [\mathbf{e}_v \mid \mathbf{e}_{\mathrm{slc}}] \in \mathbb{R}^{B \times 29 \times T},
\]
where $\mathbf{e}_v \in \mathbb{R}^{B \times 17 \times T}$ denotes the 17 observed visual-related EEG channels and $\mathbf{e}_{\mathrm{slc}} \in \mathbb{R}^{B \times 12 \times T}$ denotes the proposed 12 semantic latent channels generated from $\mathbf{e}_v$ through the learnable channel projection described in Section~\ref{subsubsec:semantic_latent_channels}. An RST-GAT encoder, similar to that used in Phase-I, is then applied to the joint channel representation:
\begin{equation}
\label{eq:phaseII-A}
\ddot{\mathbf{e}}_2 = f_{\mathrm{ST\mbox{-}GAT}}\!\left(\ddot{\mathbf{e}} \otimes \ddot{\mathbf{w}}_t + \ddot{\mathbf{e}}\right) + \ddot{\mathbf{e}},
\end{equation}
where $\otimes$ denotes element-wise multiplication with broadcasting along the channel dimension, and $\ddot{\mathbf{w}}_t \in \mathbb{R}^{1 \times T}$ denotes the normalized temporal weight vector produced by the Phase-II temporal weighting module. For notational convenience, we partition the encoded joint representation as
$\ddot{\mathbf{e}}_2 = [\ddot{\mathbf{e}}_{v} \mid \ddot{\mathbf{e}}_{\mathrm{slc}}], $
where $\ddot{\mathbf{e}}_{v}$ and $\ddot{\mathbf{e}}_{\mathrm{slc}}$ denote the encoded components corresponding to the observed visual channels and the semantic latent channels, respectively.

Based on the encoded joint representation, we construct two semantic branches.
For the \textbf{coarse semantic branch}, we use the encoded semantic-latent-channel component $\ddot{\mathbf{e}}_{\mathrm{slc}}$ to model abstract category-level semantics. Rather than collapsing the temporal dimension, we first apply temporal weighting to emphasize semantically informative time steps and then vectorize the weighted representation:
\begin{equation}
\label{eq:phaseII-coarse-agg}
\tilde{\mathbf{e}}_{\mathrm{coarse}} = \mathrm{vec}\!\left(\ddot{\mathbf{e}}_{\mathrm{slc}} \odot \ddot{\mathbf{w}}_t\right),
\end{equation}
\vspace{-0.25in}
\begin{equation}
\label{eq:phaseII-coarse}
\hat{\mathbf{e}}_{\mathrm{coarse}} = f_{\mathrm{coarse}}\!\left(\tilde{\mathbf{e}}_{\mathrm{coarse}}\right),
\end{equation}
where $\odot$ denotes element-wise temporal reweighting with broadcasting along the channel dimension, and $f_{\mathrm{coarse}}(\cdot)$ denotes the coarse semantic projection head summarized in Table~\ref{tab:full_backbone_setting}, which maps the weighted and vectorized semantic-latent-channel representation to a $1024$-dimensional embedding.
The resulting coarse representation $\hat{\mathbf{e}}_{\mathrm{coarse}} \in \mathbb{R}^{B \times 1024}$ is aligned with the text semantic representation $\mathbf{t}_{\mathrm{coarse}} \in \mathbb{R}^{B \times 1024}$ extracted from the pretrained CLIP text encoder. The corresponding alignment loss is
\begin{equation}
\label{eq:clip_II_coarse}
\mathcal{L}^{II}_{\mathrm{CLIP\mbox{-}c}} = \mathrm{CLIP\_LOSS}(\hat{\mathbf{e}}_{\mathrm{coarse}}, \mathbf{t}_{\mathrm{coarse}}).
\end{equation}

For the \textbf{fine semantic branch}, we use the full encoded joint representation $\ddot{\mathbf{e}}_2$ to capture fine image-level visual-semantic information from interactions between observed visual EEG channels and semantic latent channels. The representation is obtained by
\begin{equation}
\label{eq:phaseII_fine}
\hat{\mathbf{e}}_{\mathrm{fine}} = f_{\mathrm{fine}}\!\left(\mathrm{vec}(\ddot{\mathbf{e}}_2)\right),
\end{equation}
where $f_{\mathrm{fine}}(\cdot)$ denotes the fine semantic projection head summarized in Table~\ref{tab:full_backbone_setting}, which maps the vectorized joint representation to a $1024$-dimensional embedding $\hat{\mathbf{e}}_{\mathrm{fine}} \in \mathbb{R}^{B \times 1024}$. The embedding is then aligned with the fine image-level visual-semantic representation $\mathbf{v}_{\mathrm{fine}} \in \mathbb{R}^{B \times 1024}$ obtained from the projected \texttt{block3} output of the frozen OpenCLIP ResNet50 image encoder. The corresponding alignment loss is
\begin{equation}
\label{eq:clip_II_fine}
\mathcal{L}^{II}_{\mathrm{CLIP\mbox{-}f}} = \mathrm{CLIP\_LOSS}(\hat{\mathbf{e}}_{\mathrm{fine}}, \mathbf{v}_{\mathrm{fine}}).
\end{equation}

Through these two branches, Phase-II explicitly separates abstract label-level semantics from finer image-level semantics, while preserving their interaction within a shared joint channel representation. This design provides structured semantic supervision for the subsequent integration stage in Phase-III.

\subsection{Phase-III: Integrative Information Fusion}
\label{sec:Phase-III}

Within the proposed staged framework, \textbf{Phase-III} performs integrative information fusion by combining the low-level visual representation learned in Phase-I with the structured semantic representations learned in Phase-II. While Phase-I captures early perceptual information and Phase-II models higher-level semantics at both coarse and fine levels, Phase-III integrates these complementary representations into a unified EEG embedding for final cross-modal alignment. This design is motivated by neuroscientific evidence suggesting that higher-level cognition arises from the progressive integration of multiple information pathways~\citep{felleman1991distributed, graumann2022spatiotemporal, goodale1992separate}.

Specifically, we first combine the low-level visual representation $\mathbf{e}_1$ from Phase-I with the fine semantic representation $\hat{\mathbf{e}}_{\mathrm{fine}}$ from Phase-II, and map the concatenated feature to a unified embedding space through an MLP. This step preserves both early perceptual cues and more detailed image-level visual-semantic information. We then incorporate the coarse semantic representation $\hat{\mathbf{e}}_{\mathrm{coarse}}$ as additive category-level semantic guidance. In this way, the final EEG representation jointly reflects perceptual detail, fine semantic specificity, and coarse semantic abstraction. Formally, the integrative information fusion is defined as
\begin{equation}
\mathbf{e}_{\mathrm{EEG}} = f_{\mathrm{MLP}}([\mathbf{e}_1 \mid \hat{\mathbf{e}}_{\mathrm{fine}}]) + \hat{\mathbf{e}}_{\mathrm{coarse}},
\label{eq:phase_III_eeg}
\end{equation}
where $[\cdot | \cdot]$ denotes concatenation. The resulting representation $\mathbf{e}_{\mathrm{EEG}} \in \mathbb{R}^{B \times 1024}$ is aligned with the final image embedding $\mathbf{v}_{\mathrm{image}} \in \mathbb{R}^{B \times 1024}$ extracted from the final output of the frozen OpenCLIP ResNet50 image encoder through
\begin{equation}
\mathcal{L}^{III}_{\mathrm{CLIP}} = \mathrm{CLIP\_LOSS}(\mathbf{e}_{\mathrm{EEG}}, \mathbf{v}_{\mathrm{image}}).
\label{eq:clip_III}
\end{equation}

The overall training objective is defined as the weighted sum of the losses from the three phases:
\begin{equation}
\mathcal{L}_{\mathrm{total}} =
\alpha_1 \mathcal{L}^{I}_{\mathrm{CLIP}}
+ \alpha_2 \left(\mathcal{L}^{II}_{\mathrm{CLIP\mbox{-}c}} + \mathcal{L}^{II}_{\mathrm{CLIP\mbox{-}f}}\right)
+ \alpha_3 \mathcal{L}^{III}_{\mathrm{CLIP}},
\label{eq:loss_total}
\end{equation}
where $\alpha_1$, $\alpha_2$, and $\alpha_3$ balance the contributions of the three phases, and are set to 0.1, 0.2, and 0.5, respectively.

\section{Experiments}
\label{sec:exp}
 
\subsection{Experimental Setup}
\subsubsection{Benchmark Dataset}
The benchmark dataset THINGS-EEG \citep{gifford2022large} was employed for all experiments. THINGS-EEG is currently the largest publicly available EEG dataset for brain decoding, and it has become a widely recognised benchmark in recent top-tier conference and journal publications, particularly in the context of zero-shot learning settings. The dataset was designed to capture rich and generalisable neural representations of visual-semantic concepts, thereby providing a challenging and comprehensive testbed for evaluating model generalisation across subjects and unseen categories. 
Data were collected using a Rapid Serial Visual Presentation paradigm, in which visual stimuli were presented in rapid succession while EEG signals were recorded. The dataset contains neural responses from 10 participants, each exposed to a broad spectrum of object categories encompassing diverse visual and semantic domains.
The \textbf{training set} consists of 1,654 object classes, each represented by 10 different images, with each image presented four times in a randomised sequence. This yields a total of 66,160 EEG samples. The \textbf{test set} contains 200 held-out classes, each represented by a single image repeated 80 times, resulting in 16,000 EEG samples. All stimuli were presented in a randomised order to reduce habituation and expectancy effects. In addition to the standard benchmark setting, we further consider an expanded multi-image evaluation protocol based on the same 200 unseen classes to provide a more comprehensive assessment of fine-grained retrieval, as detailed in Section~\ref{subsec:Expanded experiment}.

We evaluate the proposed method under two standard settings.
\begin{itemize}[itemsep=0pt, topsep=0pt]
    \item \textbf{Subject-dependent 200-way zero-shot classification}. The model is trained on the training set and evaluated on the test set of the same subject. 
    \item \textbf{Subject-independent 200-way zero-shot classification}. The model is trained entirely on the training sets of other subjects and evaluated on the test set of the target subject.
\end{itemize}

\subsubsection{Baselines and Evaluation Metrics}
We compare our method with nine representative EEG visual decoding approaches evaluated on the same THINGS-EEG benchmark, including \citet{wu2025bridging} (CVPR'25), \citet{zhang2025cognitioncapturer} (AAAI'25), \citet{jing2025pinpointing} (KBS'25), \citet{zhang2025category} (IJCAI'25), \citet{shi2025brainalign} (MICCAI'25), \citet{li2024visual} (NeurIPS'24), \citet{song2024decoding} (ICLR'24), \citet{chen2024visual} (arXiv'24), and \citet{du2023decoding} (TPAMI'23). These methods cover several representative directions in EEG visual decoding, including cross-modal alignment, semantic regularization, wavelet-domain modeling, diffusion-based decoding, graph-based modeling, and vision-language supervision. For fair comparison, all baseline results are taken from the corresponding papers or official released results reported under the standard THINGS-EEG 200-way zero-shot protocol. 

Following the standard evaluation protocol used in prior work, we report Top-1 and Top-5 accuracy under both subject-dependent and subject-independent 200-way zero-shot settings. These metrics provide a consistent basis for comparing decoding performance across methods on the THINGS-EEG benchmark.

\subsubsection{Key Implementation Details}
The detailed architectural specification of the proposed framework is summarized in Table~\ref{tab:full_backbone_setting}. All experiments were carried out on a single NVIDIA GeForce RTX 4090 GPU, with the framework implemented in PyTorch. For EEG signal preprocessing, we follow the standard pipeline described in \cite{song2024decoding, wu2025bridging}. For visual feature extraction, we employ the pretrained OpenCLIP ResNet50 image encoder~\citep{ilharco_gabriel_2021_5143773}, keeping its parameters fixed throughout training. The model is optimized using AdamW with a learning rate of $1\times10^{-4}$ and a batch size of 1024. Training is conducted for 40 epochs, and the random seed is fixed to 42. Unless otherwise stated, the same hyperparameter configuration is used for both subject-dependent and subject-independent experiments.




\begin{table*}[!bt]
\centering
\caption{Architecture and hyper-parameters of the proposed neuroscience-inspired staged EEG representation learning framework.}
\label{tab:full_backbone_setting}
\small
\setlength{\tabcolsep}{5pt}
\renewcommand{\arraystretch}{1.12}
\begin{tabular}{p{0.22\linewidth} p{0.76\linewidth}}
\toprule
\textbf{Module} & \textbf{Setting} \\
\midrule

EEG input &
Phase-I uses visual EEG
$\ve_v \in \mathbb{R}^{B \times 17 \times T}$;
Phase-II uses expanded EEG
$\ddot{\ve} = [\ve_v \mid \ve_{\mathrm{slc}}] \in \mathbb{R}^{B \times 29 \times T}$, with $T=175$ \\

Channel partition &
$17$ visual-related EEG channels + $12$ semantic latent channels \\

Visual-related channels &
$\{$P7, P5, P3, P1, Pz, P2, P4, P6, P8, PO7, PO3, POz, PO4, PO8, O1, Oz, O2$\}$ \\

Backbone structure &
Three-phase RST-GAT framework consisting of
Phase-I low-level visual representation learning,
Phase-II dual-level semantic representation learning,
and Phase-III integrative information fusion \\

Semantic latent channel &
\texttt{SLCInitLinear}: Linear$(17 \rightarrow 12)$ applied at each time step to generate
$\ve_{\mathrm{slc}} \in \mathbb{R}^{B \times 12 \times T}$ from visual EEG \\

Phase-I input &
$\ve_v \in \mathbb{R}^{B \times 17 \times 175}$ \\

Phase-II input &
$\ddot{\ve} = [\ve_v \mid \ve_{\mathrm{slc}}] \in \mathbb{R}^{B \times 29 \times 175}$ \\

Adaptive temporal weighting &
One learnable temporal weighting module for Phase-I and one independent module for Phase-II:
Linear$(T \rightarrow 128)$ + ReLU + Linear$(128 \rightarrow T)$ + Softmax \\

\midrule

Channel graph (Phase-I) &
Fully connected directed graph over $17$ visual channels, i.e., all pairs $(i,j)$ with $i \neq j$ \\

Channel graph (Phase-II) &
Fully connected directed graph over $29$ expanded EEG channels, i.e., all pairs $(i,j)$ with $i \neq j$ \\

Temporal graph &
Fully connected directed graph over temporal indices $t \in [0, T-1]$, i.e., all pairs $(i,j)$ with $i \neq j$ \\

GNN backbone &
\texttt{GATConv} from \texttt{torch\_geometric.nn} \\

Phase-I channel GAT &
\texttt{GATConv}: in$=T$, out$=T$, heads$=1$, concat$=\mathrm{False}$ \\

Phase-I temporal GAT &
\texttt{GATConv}: in$=17$, out$=17$, heads$=1$, concat$=\mathrm{False}$ \\

Phase-II channel GAT &
\texttt{GATConv}: in$=T$, out$=T$, heads$=1$, concat$=\mathrm{False}$ \\

Phase-II temporal GAT &
\texttt{GATConv}: in$=29$, out$=29$, heads$=1$, concat$=\mathrm{False}$ \\

Residual update &
For each phase, $x' = x + \mathrm{GAT}(\mathrm{Attn}(x))$ \\

\midrule

Phase-I encoder &
Flatten + Linear$(17 \cdot 175 \rightarrow 256)$ + GELU + LayerNorm,
producing low-level EEG embedding
$\ve_1 \in \mathbb{R}^{B \times 256}$ \\

Phase-II fine encoder &
Flatten + Linear$(29 \cdot 175 \rightarrow 1024)$ + GELU + LayerNorm,
producing fine semantic embedding
$\hat{\ve}_{\mathrm{fine}} \in \mathbb{R}^{B \times 1024}$ \\

Phase-II coarse encoder &
Attention-weighted flatten over refined semantic-latent-channel features
+ Linear$(12 \cdot 175 \rightarrow 1024)$
+ Residual(GELU + Linear + Dropout$(0.1)$)
+ LayerNorm,
producing coarse semantic embedding
$\hat{\ve}_{\mathrm{coarse}} \in \mathbb{R}^{B \times 1024}$ \\

Phase-III information fusion &
Concatenate $[\ve_1 \mid \hat{\ve}_{\mathrm{fine}}]$
+ Linear$(256 + 1024 \rightarrow 1024)$ \\

Output embedding &
Final EEG embedding
$\ve_{\mathrm{EEG}} = f_{\mathrm{MLP}}([\ve_1 \mid \hat{\ve}_{\mathrm{fine}}]) + \hat{\ve}_{\mathrm{coarse}}
\in \mathbb{R}^{B \times 1024}$ \\

\midrule

Low-level image feature &
\texttt{block1} output of the frozen OpenCLIP ResNet50 image encoder projected to $256$-d by an MLP \\

High-level image feature &
\texttt{block3} output of the frozen OpenCLIP ResNet50 image encoder projected to $1024$-d by an MLP \\

Text feature &
OpenCLIP text encoder output for class labels,
$\mathbf{t}_{\mathrm{coarse}} \in \mathbb{R}^{B \times 1024}$ \\

Final image embedding &
Final output of the frozen OpenCLIP ResNet50 image encoder,
$\mathbf{v}_{\mathrm{image}} \in \mathbb{R}^{B \times 1024}$, for Phase-III alignment \\

Image feature projector 1 &
Linear$(256 \rightarrow 256)$ + ReLU + Linear$(256 \rightarrow 256)$ \\

Image feature projector 3 &
Linear$(1024 \rightarrow 1024)$ + ReLU + Linear$(1024 \rightarrow 1024)$ \\

Image feature normalization &
$L_2$ normalization on projected image features \\

Temperature &
Learnable logit scale initialized as $\log(1/0.07)$; \texttt{Softplus} applied \\

\bottomrule
\end{tabular}
\end{table*}

\subsection{Main Results}
\subsubsection{Subject-Dependent Zero-Shot Evaluation}
Table~\ref{tab:sub-dependent} presents the 200-way zero-shot classification results under the subject-dependent setting. Our method achieves the best Top-1 accuracy on all 10 subjects, with an average of \textbf{55.0\%}, surpassing the strongest prior baseline, \citet{wu2025bridging} (CVPR), by \textbf{4.1} absolute points over its 50.9\% average, corresponding to a relative improvement of \textbf{8.1\%}. The gains are consistent across subjects, with particularly large margins on Sub01 (+7.3) and Sub09 (+7.4), indicating improved robustness on more challenging participants. Compared with recent alignment-based models, including \citet{chen2024visual} (arXiv) (37.2\%), \citet{shi2025brainalign} (MICCAI) (30.6\%), \citet{zhang2025category} (IJCAI) (33.4\%), \citet{zhang2025cognitioncapturer} (AAAI) (35.6\%), and \citet{li2024visual} (NeurIPS) (27.1\%), our approach improves the average Top-1 accuracy by \textbf{17.8} to \textbf{26.5} percentage points, highlighting the advantage of staged learning and channel-level modeling over conventional global embedding alignment.

For Top-5 accuracy, our method again ranks first, achieving an average of \textbf{84.2\%}. This exceeds \citet{zhang2025cognitioncapturer} (AAAI) at 80.2\% and \citet{wu2025bridging} (CVPR) at 79.7\% by \textbf{4.0} and \textbf{4.5} percentage points, respectively. Our method achieves the best Top-5 performance on 8 out of 10 subjects and remains competitive on the remaining two subjects, namely Sub01 and Sub04, while consistently outperforming \citet{wu2025bridging} (CVPR) across all subjects.
\begin{table*}[h]
    \centering
    \caption{Subject-dependent Top-1 and Top-5 accuracy (\%) in 200-way zero-shot classification.}
    \label{tab:sub-dependent}
    \setlength{\tabcolsep}{4pt}
    \renewcommand{\arraystretch}{1.1}
    \resizebox{\textwidth}{!}
    {%
        \begin{tabular}{llccccccccccc}
            \toprule
            \textbf{Metrics} & \textbf{Methods} & \textbf{Sub01} & \textbf{Sub02} & \textbf{Sub03} & \textbf{Sub04} & \textbf{Sub05} & \textbf{Sub06} & \textbf{Sub07} & \textbf{Sub08} & \textbf{Sub09} & \textbf{Sub10} & \textbf{Avg.}\\
            \midrule
            \multirow{10}{*}{\makecell{Top-1\\Accuracy}}
			& \citet{du2023decoding} (TPAMI)                    & 6.1  & 4.9  & 5.6  & 5.0  & 4.0  & 6.0  & 6.5  & 8.8  & 4.3  & 7.0  & 5.8 \\
            &\citet{song2024decoding} (ICLR)                   & 13.2 & 13.5 & 14.5 & 20.6 & 10.1 & 16.5 & 17.0 & 22.9 & 15.4 & 17.4 & 16.1 \\
            &\citet{li2024visual} (NeurIPS)                    & 25.6 & 22.0 & 25.0 & 31.4 & 12.9 & 21.3 & 30.5 & 38.8 & 34.4 & 29.1 & 27.1 \\
            &\citet{chen2024visual} (arXiv)                    & 32.6 & 34.4 & 38.7 & 39.8 & 29.4 & 34.5 & 34.5 & 49.3 & 39.0 & 39.8 & 37.2 \\
            &\citet{shi2025brainalign}(MICCAI)
            & 32.0 & 25.5 & 34.0 & 37.0 & 22.0 & 27.0 & 28.5 & 43.0 & 26.5 & 30.5 & 30.6\\
            &\citet{zhang2025category} (IJCAI)                 & 33.0 & 28.0 & 33.5 & 36.0 & 26.0 & 30.5 & 34.0 & 43.0 & 31.5 & 38.5 & 33.4 \\
            &\citet{zhang2025cognitioncapturer} (AAAI)         & 31.4 & 31.4 & 38.2 & 40.4 & 24.4 & 34.8 & 34.7 & 48.1 & 37.4 & 35.6 & 35.6 \\
            &\citet{jing2025pinpointing} (KBS)                 &
            32.6 & 32.5 & 35.0 & 39.8 & 26.5 & 34.7 & 36.5 & 48.3 & 37.8 & 36.6 & 36.0\\
            &\citet{wu2025bridging} (CVPR)                     &  {41.2} &  {51.2} &  {51.2} &  {51.1} &  {42.2} &  {57.5} &  {49.0} &  {58.6} &  {45.1} &  {61.5} &  {50.9} \\
            & Ours                                               & \textbf{48.5} & \textbf{56.0} & \textbf{53.5} & \textbf{54.0} & \textbf{44.0} & \textbf{60.0} & \textbf{51.5} & \textbf{64.0} & \textbf{52.5} & \textbf{66.0} & \textbf{55.0} \\
            \midrule
            \midrule
            \multirow{10}{*}{\makecell{Top-5\\Accuracy}}
            &\citet{du2023decoding} (TPAMI)                    & 17.9 & 14.9 & 17.4 & 15.1 & 13.4 & 18.2 & 20.4 & 23.7 & 14.0 & 19.7 & 17.5 \\
            &\citet{song2024decoding} (ICLR)                   & 39.5 & 40.3 & 42.7 & 52.7 & 31.5 & 44.0 & 42.1 & 56.1 & 41.6 & 45.8 & 43.6 \\
            &\citet{li2024visual} (NeurIPS)                    & 60.4 & 54.5 & 62.4 & 60.9 & 43.0 & 51.1 & 61.5 & 72.0 & 51.5 & 63.5 & 58.1 \\
            &\citet{chen2024visual} (arXiv)                    & 63.7 & 69.9 & 73.5 & 72.0 & 58.6 & 68.8 & 68.3 & 79.8 & 69.6 & 75.3 & 69.9 \\
            &\citet{shi2025brainalign}(MICCAI) 				   & 63.0 & 59.0 & 69.0 & 57.0 & 46.0 & 47.5 & 59.5 & 74.5 & 57.5 & 66.0 & 59.9\\
            &\citet{zhang2025category} (IJCAI)                 & 58.5 & 56.5 & 61.0 & 68.0 & 48.0 & 62.5 & 62.5 & 73.5 & 58.5 & 69.0 & 61.8 \\
            &\citet{zhang2025cognitioncapturer} (AAAI)         & \textbf{79.7} & 77.8 &  {85.7} & \textbf{85.8} & 66.3 & 78.8 &  {81.0} &  {88.6} &  {79.4} & 79.3 &  {80.2} \\
            &\citet{jing2025pinpointing} (KBS)                 & 77.9 & 78.9 & 85.4 & 85.1 & 73.4 & 78.2 & 80.4 & 83.7 & 84.0 & 80.7 & 80.8            \\
            &\citet{wu2025bridging} (CVPR)                     & 70.5 &  {80.9} & 82.0 & 76.9 &  {72.8} &  {83.5} & 79.9 & 85.8 & 76.2 &  {88.2} & 79.7 \\
            & Ours                       &  {74.0} & \textbf{87.5} & \textbf{88.0} &  {80.0} & \textbf{79.5} & \textbf{88.0} & \textbf{83.0} & \textbf{89.0} & \textbf{81.5} & \textbf{91.0} & \textbf{84.2} \\
            \bottomrule
        \end{tabular}%
    }
\end{table*}
\begin{table*}[t]
    \centering
    \caption{Subject-independent Top-1 and Top-5 accuracy (\%) in 200-way zero-shot classification. (Note that some methods in Table~\ref{tab:sub-dependent} were designed exclusively for the subject-dependent setting. ``--'' indicates that subject-wise results were not reported in the corresponding paper.)}
    \label{tab:sub-independent}
    \setlength{\tabcolsep}{4pt}
    \renewcommand{\arraystretch}{1.1}
    \resizebox{\textwidth}{!}
    {%
        \begin{tabular}{llccccccccccc}
            \toprule
            \textbf{Metrics} &\textbf{Methods} & \textbf{Sub01} & \textbf{Sub02} & \textbf{Sub03} & \textbf{Sub04} & \textbf{Sub05} & \textbf{Sub06} & \textbf{Sub07} & \textbf{Sub08} & \textbf{Sub09} & \textbf{Sub10} & \textbf{Avg.}\\
            \midrule
            \multirow{6}{*}{\makecell{Top-1\\Accuracy}} 
			& \citet{du2023decoding} (TPAMI)      &2.3  &1.5  &1.4 &1.7 &1.5 &1.8 &2.1 &2.2 &1.6  &2.3 &1.8\\
            & \citet{song2024decoding} (ICLR)     &7.6  &5.9  &6.0 &6.3 & 4.4 &5.6 &5.6 &6.3 &5.7  &8.4 &6.2\\
            & \citet{li2024visual} (NeurIPS)      &10.5 &7.1  &  \textbf{11.9}&  \textbf{14.7}&7.0 &11.1&  \textbf{16.1}&  \textbf{15.0}&4.9  & \textbf{20.5}&11.8\\ 
            &\citet{shi2025brainalign}(MICCAI)
            & -- & -- & -- & -- & -- & -- & -- & -- & -- & -- & 12.4\\
            & \citet{wu2025bridging} (CVPR)       & {11.5} & {15.5} & {9.8} &13.0& {8.8} & {11.7}& {10.2}& {12.2} &  \textbf{15.5} & {16.0}& {12.4} \\
            & Ours                                &  \textbf{13.0} &  \textbf{16.5} &8.0 & {14.5}&  \textbf{10.0}&  \textbf{14.0}&9.5&11.5& {14.5} &  \textbf{20.5}&  \textbf{13.2} \\
            \midrule
            \midrule
			\multirow{6}{*}{\makecell{Top-5\\Accuracy}} 
            & \citet{du2023decoding} (TPAMI)      &8.0  &6.3  &5.9  &6.7 &5.6 &7.2 &8.1 &7.6  &6.4 &8.5  &7.0\\
            & \citet{song2024decoding} (ICLR)     &22.8 &20.5 &22.3 &20.7&18.3&22.2&19.7&22.0 &17.6&28.3 &21.4\\
            & \citet{li2024visual} (NeurIPS)      &26.8 &24.8 &  \textbf{33.8} &  \textbf{39.4}&23.9&  \textbf{35.8}&  \textbf{43.5}&  \textbf{40.3} &22.7&  \textbf{46.5} &  \textbf{33.7}\\
            & \citet{shi2025brainalign}(MICCAI)
            & -- & -- & -- & -- & -- & -- & -- & -- & -- & -- & 30.3\\
            & \citet{wu2025bridging} (CVPR)       & {29.7} & {40.0} & {27.0} &32.3&  \textbf{33.8}&31.0&23.8& {32.2} &  \textbf{40.5}& {43.5} & {33.4}\\
            
            & Ours                                &  \textbf{32.0} &  \textbf{41.5} &22.0 & {34.5}& {31.5}& {31.5}& {27.0}&30.5 & {32.0} &40.0 &32.3\\
            \bottomrule
        \end{tabular}%
    }
\end{table*}

\subsubsection{Subject-Independent Zero-Shot Evaluation}
Table~\ref{tab:sub-independent} presents the 200-way zero-shot classification results under the subject-independent setting. Under this more challenging evaluation protocol, our method achieves the best average Top-1 accuracy of 13.2\%, outperforming the strongest prior baseline, \citet{wu2025bridging} (CVPR), which achieves 12.4\%, by 0.8 percentage points on average. Notably, our framework outperforms all prior methods on several subjects, including Sub01, Sub02, Sub05, Sub06, and Sub10, and remains on par with the strongest baseline on Sub04 and Sub09. Compared with \citet{li2024visual} (NeurIPS) at 11.8\% and \citet{song2024decoding} (ICLR) at 6.2\%, our method improves the average Top-1 accuracy by 1.4 and 7.0 percentage points, respectively. These results suggest that the proposed staged learning framework and channel-level augmentation strategy improve exact cross-subject retrieval performance.
For Top-5 accuracy, our method achieves an average of 32.3\%, which remains competitive with \citet{li2024visual} (NeurIPS) at 33.7\% and \citet{wu2025bridging} (CVPR) at 33.4\%, while substantially outperforming \citet{song2024decoding} (ICLR) at 21.4\% and \citet{du2023decoding} (TPAMI) at 7.0\%. Overall, these results indicate that the proposed method provides the strongest exact retrieval performance under subject-independent evaluation while maintaining competitive broader retrieval capability.

\subsubsection{Discussion on the Standard Benchmark}
A clear performance gap remains between the subject-dependent and subject-independent settings, which is consistent with the well-recognized challenge of cross-subject EEG generalization. EEG signals are highly subject-specific and are influenced by multiple factors, including individual differences in brain anatomy, electrode placement, and neural response patterns. As a result, subject-independent decoding is substantially more difficult than within-subject decoding.
Against this background, the proposed framework exhibits two notable characteristics. First, it achieves the strongest average Top-1 accuracy under both evaluation settings, indicating that the staged representation learning strategy is effective for learning discriminative EEG representations. Second, although performance decreases under the more challenging subject-independent setting, the proposed method still maintains competitive Top-5 retrieval performance while preserving the best average Top-1 accuracy. These results suggest that explicitly modeling low-level perception, high-level semantics, and integrative fusion provides a useful representation framework for both within-subject decoding and cross-subject generalization.

\subsection{Additional Analyses}
\label{sec:additional_analyses}

In addition to the main results, we conduct several additional analyses under the standard 200-way zero-shot protocol to better understand the behavior of the proposed framework. We examine the retrieval performance of representations learned at different stages, analyze the contribution of the Phase-II semantic branches, provide qualitative retrieval examples, and investigate how discriminative information accumulates over time. These analyses are intended to complement the main quantitative comparison and provide further evidence for the effectiveness of the staged decomposition and structured semantic modeling strategy.
\subsubsection{Layer-Wise Retrieval Analysis}

\begin{table*}[h]
    \centering
    \caption{Layer-wise retrieval accuracy (\%) of different representation levels in subject-dependent 200-way zero-shot evaluation.}
    \label{tab:hierarchical-analysis}
    \setlength{\tabcolsep}{5pt}
    \renewcommand{\arraystretch}{1.1}
    \resizebox{\textwidth}{!}
    {%
        \begin{tabular}{llccccccccccc}
            \toprule
            \textbf{Metrics}& \textbf{Representations} & \textbf{Sub01} & \textbf{Sub02} & \textbf{Sub03} & \textbf{Sub04} & \textbf{Sub05} & \textbf{Sub06} & \textbf{Sub07} & \textbf{Sub08} & \textbf{Sub09} & \textbf{Sub10} & \textbf{Avg.}\\
            \midrule
			\multirow{3}{*}{\makecell{Top-1\\Accuracy}} 
            & Phase-I & 30.0 & 28.5 & 25.0 & 24.0 & 23.5 & 35.0 & 27.5 & 31.0 & 25.0 & 30.0 & 27.9 \\
            & Phase-II(fine)  & 49.5 & 47.5 & 50.5 & 49.5 & 43.0 & 47.5 & 45.5 & 51.0 & 43.0 & 44.0 & 47.1 \\
            & Phase-III        & 48.5 & 56.0 & 53.5 & 54.0 & 44.0 & 60.0 & 51.5 & 64.0 & 52.5 & 66.0 & 55.0 \\
            \midrule
            \midrule
			\multirow{3}{*}{\makecell{Top-5\\Accuracy}} 
            & Phase-I & 62.0 & 58.0 & 61.0 & 54.5 & 47.0 & 67.5 & 55.5 & 65.0 & 57.5 & 65.0 & 59.3 \\
            & Phase-II(fine)  & 77.0 & 80.5 & 79.5 & 83.5 & 75.5 & 84.5 & 76.5 & 81.5 & 74.5 & 82.5 & 79.5 \\
            & Phase-III        & 74.0 & 87.5 & 88.0 & 80.0 & 79.5 & 88.0 & 83.0 & 89.0 & 81.5 & 91.0 & 84.2 \\
            \bottomrule
        \end{tabular}%
    }
\end{table*}

Table~\ref{tab:hierarchical-analysis} provides a layer-wise evaluation of the representations learned at different stages of the proposed framework. The Phase-I representation, which mainly captures low-level perceptual information, achieves the lowest average retrieval accuracy, with 27.9\% Top-1 and 59.3\% Top-5. When the model progresses to Phase-II, the fine semantic representation improves the average performance to 47.1\% Top-1 and 79.5\% Top-5, indicating that high-level semantic learning substantially enhances discriminative capability beyond shallow visual similarity. The final fused representation in Phase-III further improves the average performance to 55.0\% Top-1 and 84.2\% Top-5. 
Although the improvement is not strictly monotonic for every individual subject and metric, the average results show a clear hierarchical progression from Phase-I to Phase-II and then to Phase-III. This trend provides quantitative evidence that the proposed framework learns a hierarchy of complementary EEG representations rather than relying on a single undifferentiated embedding space. In particular, the results suggest that low-level perceptual representation, fine-grained semantic modeling, and their subsequent integration with coarse semantic information make distinct contributions, and that their fusion yields the strongest overall retrieval performance.

\subsubsection{Phase-II Semantic Branch Analysis}
\label{sec:semantic_alignment_analysis}


To further evaluate the proposed coarse semantic branch, we analyze the alignment between the coarse EEG semantic representations learned from the 12 semantic latent channels and the corresponding text embeddings. Specifically, the features derived from the semantic latent channels are projected by the coarse semantic branch to obtain $\hat{\mathbf{e}}_{\mathrm{coarse}}$, and retrieval is then performed against the target text embeddings extracted from the pretrained CLIP text encoder.
Across the 10 subjects, the projected embeddings from the semantic latent channels achieve an average Top-1 / Top-5 text retrieval accuracy of 10.35\% / 29.55\%. Although these values remain substantially lower than the final image retrieval results, they nevertheless indicate that the coarse branch captures non-trivial label-level semantic information directly from EEG signals. This result suggests that the semantic latent channels do more than enlarge the feature space, and instead provide a useful representation basis for coarse semantic abstraction.

This interpretation is further supported by the ablation results in Table~\ref{tab:sub-dependent-ab}. When the proposed semantic latent channels are replaced with original language-related channels, or when the coarse semantic branch is removed, the overall decoding performance decreases. Taken together, these results suggest that the coarse semantic branch provides useful complementary semantic supervision in Phase-II, while the proposed semantic latent channels offer a more suitable channel-level representation space for such modeling.

\begin{figure*}[t]
    \centering
    \includegraphics[width=\textwidth]{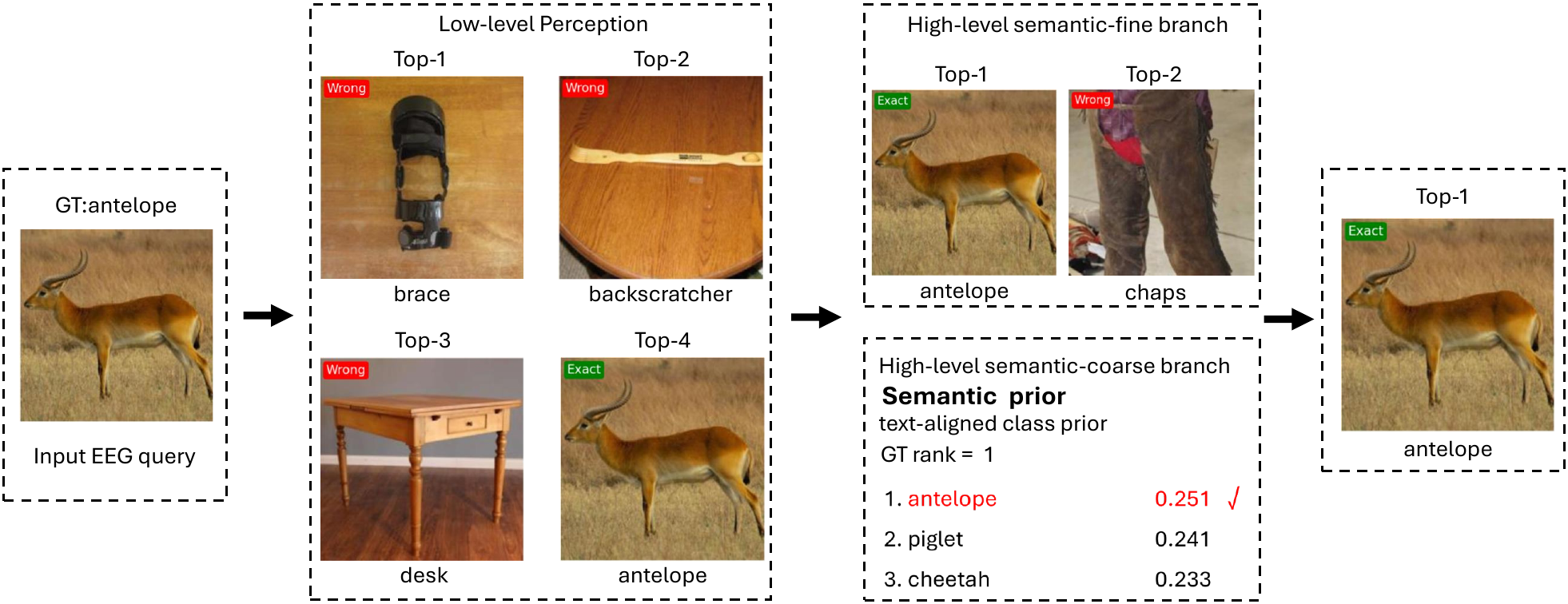}
    \caption{A branch-wise qualitative example illustrating the complementary roles of the fine semantic branch and the coarse semantic branch in the proposed staged framework. For the EEG query corresponding to the class \emph{antelope}, the low-level perception stage retrieves visually similar but semantically ambiguous candidates, while the fine semantic branch ranks the ground-truth image at Top-1. Meanwhile, the coarse semantic branch provides a text-aligned class prior that ranks the target class \emph{antelope} at the top of the semantic ranking. In the final fusion stage, the low-level representation is integrated with the fine and coarse semantic representations, yielding the correct Top-1 retrieval result. This example highlights the complementary contribution of fine image-level visual-semantic discrimination and coarse category-level guidance.}
    \label{fig:appendix-phase2-branch-example}
\end{figure*}

To further illustrate the behavior of Phase-II, Fig.~\ref{fig:appendix-phase2-branch-example} provides a branch-wise qualitative example. The example shows that the fine semantic branch mainly contributes to instance-level discrimination by promoting visually and semantically relevant candidates, whereas the coarse semantic branch provides category-level semantic guidance through text alignment. In the final fusion stage, the low-level representation is integrated with the fine and coarse semantic representations, leading to a more reliable Top-1 prediction and highlighting the complementary roles of fine image-level visual-semantic discrimination and coarse category-level semantic support.

\subsubsection{Qualitative Retrieval Examples}
\begin{figure*}[t]
    \centering
    \includegraphics[width=0.48\textwidth]{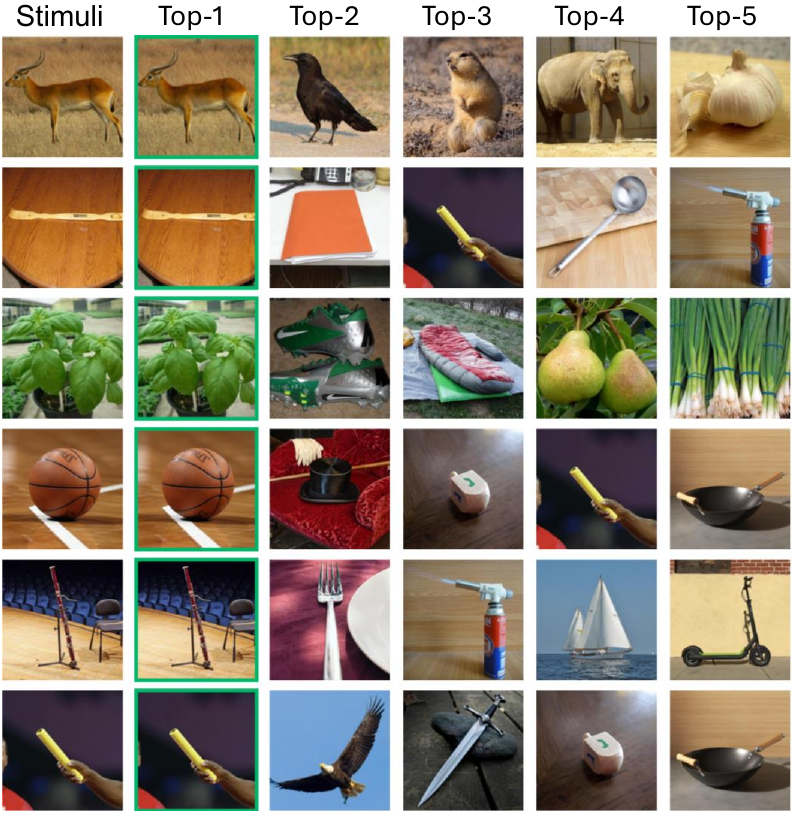}
    \hfill
    \includegraphics[width=0.48\textwidth]{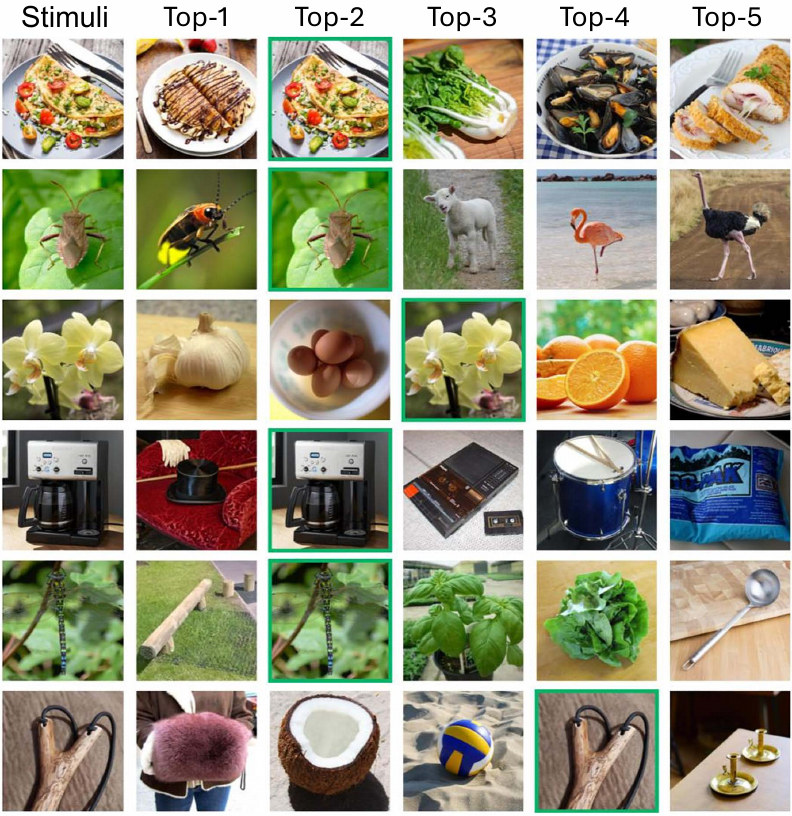}
    \caption{Qualitative retrieval examples on the standard 200-way THINGS-EEG test set, including top-1 successful cases (left) and challenging cases where the ground-truth image appears within the top-5 results (right). For each row, the query stimulus is shown in the leftmost column, followed by the top-5 retrieved images from left to right. The green box marks the ground-truth image among the retrieved results.}
    \label{fig:retrieval_cases}
\end{figure*}
We further provide qualitative retrieval examples in Fig.~\ref{fig:retrieval_cases} to better understand the retrieval behavior of the proposed framework. In successful cases, the correct stimulus image is retrieved at Top-1, indicating that the learned EEG representation can be effectively aligned with the corresponding visual embedding. These examples suggest that the proposed framework captures sufficiently discriminative concept-level information to support accurate cross-modal retrieval.

We also analyze more challenging cases in which the ground-truth image is not retrieved at Top-1 but still appears within the Top-5 results. Such cases remain informative because they indicate that the learned EEG representation often preserves substantial semantic relevance even when the exact ranking is not optimal. In other words, the failure is frequently not a complete semantic mismatch, but rather an instance-level ranking ambiguity among visually or semantically related candidates. As shown in Fig.~\ref{fig:retrieval_cases}, these ranking errors are often associated with similarities in global shape, texture, object parts, or overall category-level appearance. This observation suggests that the proposed framework already captures meaningful semantic structure, while the remaining errors mainly reflect the continuing difficulty of fine-grained instance discrimination under the standard 200-way zero-shot setting.

\begin{figure}
    \centering
    \includegraphics[width=0.99\columnwidth]{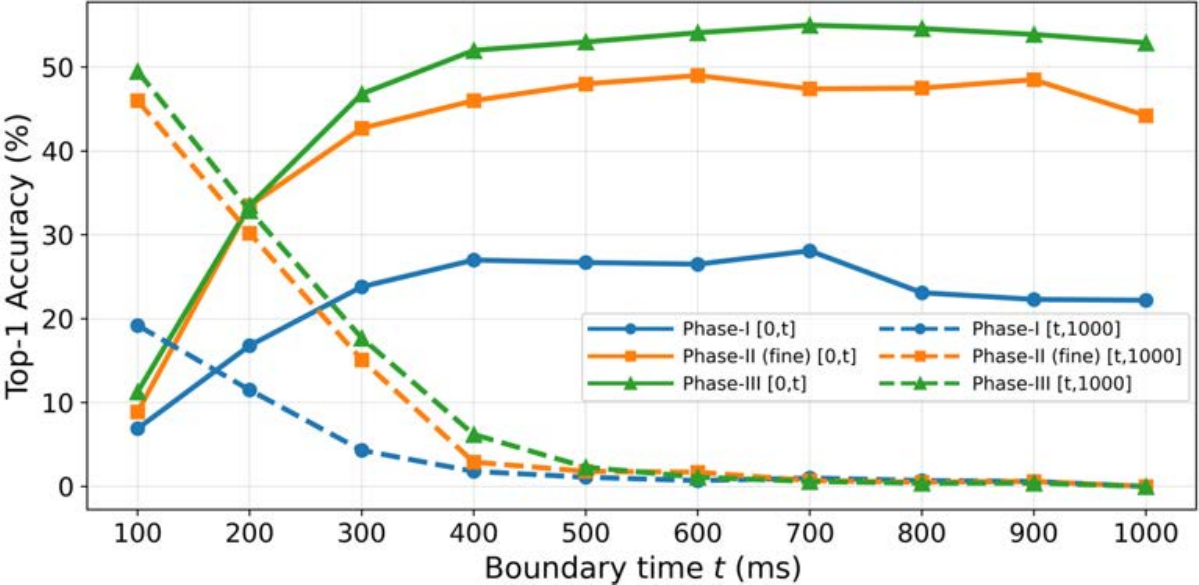}
    \caption{Temporal accumulation analysis of retrieval performance under the standard 200-way zero-shot setting. Solid lines denote Top-1 accuracy computed from EEG signals accumulated from 0 to $t$, while dashed lines denote accuracy computed from the complementary interval from $t$ to 1000 ms. Results are shown for the Phase-I low-level representation, the Phase-II fine semantic representation, and the Phase-III fused representation. When $t$ is small, the complementary interval still covers most of the trial duration, whereas its length progressively shrinks as $t$ increases. The results suggest that discriminative information is mainly accumulated in the early-to-middle period and is more effectively exploited by the semantic and fused representations.}
    \label{fig:Temporal_Accumulation_curve}
\end{figure}

\subsubsection{Temporal Accumulation Analysis}

Fig.~\ref{fig:Temporal_Accumulation_curve} presents the cumulative temporal-window analysis from 0 to 1000 ms for the Phase-I, Phase-II, and Phase-III representations. Several important observations can be drawn from the results. First, higher-level representations consistently outperform lower-level ones across nearly all temporal ranges, with the Phase-III representation achieving the strongest overall performance, followed by Phase-II, while Phase-I remains the weakest. This indicates that deeper and more structured representations are more effective at converting accumulated temporal evidence into discriminative EEG embeddings.

Second, retrieval performance increases rapidly as the temporal boundary expands from the early stage to the intermediate stage, particularly before approximately 300--400 ms. This trend suggests that a substantial portion of task-relevant information is accumulated during the early-to-middle period of the EEG response. After this range, the performance gains become much smaller and gradually approach a plateau, indicating that extending the temporal window further provides only limited additional benefit.

Third, the complementary interval analysis in Fig.~\ref{fig:Temporal_Accumulation_curve} further supports this interpretation. When only the later interval $[t,1000]$ is retained, retrieval performance generally decreases as $t$ increases and the available temporal range shrinks, suggesting that discriminative information is not uniformly distributed across the trial. Instead, the results indicate that the early-to-middle temporal period contains the most informative evidence, while later temporal segments provide more limited additional contribution when considered in isolation.

Overall, these results are consistent with the rationale of the proposed staged framework. They suggest that low-level perceptual information emerges early, while higher-level semantic and integrated representations are better able to exploit the accumulated temporal evidence over the course of the EEG response, leading to stronger retrieval performance.

\subsection{Expanded Zero-Shot Retrieval Evaluation}
\label{subsec:Expanded experiment}

\subsubsection{Expanded Multi-image Evaluation Protocol}
While the conventional 200-way zero-shot protocol provides a standard benchmark for fair comparison, its gallery contains only one image per unseen class and therefore mainly reflects coarse class-level discrimination. Such a setting is insufficient to fully evaluate our hierarchical decoding framework, which explicitly models the progressive transformation of EEG signals from early visual representations to higher-level semantic representations. To provide a more comprehensive assessment, we further construct an expanded multi-image evaluation protocol based on the same 200 unseen test classes used in prior work.

Specifically, for each of the 200 held-out classes, we retain the original test image and additionally sample 11 more images from the corresponding THINGS image database using a fixed random seed of 42, resulting in an expanded gallery of 2,400 images (200 classes $\times$ 12 images per class). Importantly, the EEG queries remain unchanged: we still use the original EEG test samples corresponding to the original 200 test images as queries, and evaluate whether the learned representations can retrieve relevant targets from the enlarged image space. In this way, the task becomes substantially more challenging, as the model must not only distinguish the correct class from 199 distractor classes, but also resolve ambiguity among multiple visual instances within the same class.

Based on this expanded gallery, we evaluate retrieval performance at multiple representation levels, including the low-level branch (\textbf{Phase-I}), the high-level fine-grained branch (\textbf{Phase-II}), and the final fused representation (\textbf{Phase-III}). We report both \textbf{category-level retrieval} and \textbf{global exact-image retrieval}. The former measures whether the retrieved results contain images from the target class in the presence of distractors from all other classes, while the latter further requires the retrieval of the exact target image from the enlarged gallery. This protocol therefore enables a more fine-grained evaluation of both category discrimination and instance-level matching.

\subsubsection{Quantitative Results on the Expanded Gallery}
\begin{table*}[ht]
    \centering
    \caption{Expanded multi-image zero-shot retrieval results (\%) on the 200 $\times$ 12 test gallery. 
    We report category-level retrieval accuracy and global exact-image retrieval accuracy for Phase-I, Phase-II, and Phase-III representations.}
    \label{tab:expanded-multi-image}
    \renewcommand{\arraystretch}{1.1}
    \resizebox{\textwidth}{!}
    {%
        \begin{tabular}{llccccccccccc}
            \toprule
            \textbf{Metrics}&\textbf{Stages} & \textbf{Sub01} & \textbf{Sub02} & \textbf{Sub03} & \textbf{Sub04} & \textbf{Sub05} & \textbf{Sub06} & \textbf{Sub07} & \textbf{Sub08} & \textbf{Sub09} & \textbf{Sub10} & \textbf{Avg.} \\
            \midrule
            \multirow{3}{*}{\makecell{Category-level\\retrieval Top-1}} 
			& Phase-I & 1.0 & 1.5 & 0.5 & 1.5 & 0.0 & 0.5 & 0.5 & 1.5 & 1.5 & 1.5 & 1.0 \\
            & Phase-II & 6.5 & 5.0 & 8.5 & 10.5 & 5.0 & 8.0 & 6.5 & 11.5 & 10.0 & 5.5 & 7.7 \\
            & Phase-III  & 9.0 & 13.0 & 18.0 & 20.5 & 11.5 & 22.0 & 17.5 & 22.5 & 15.0 & 21.5 & 17.1 \\
            \midrule
			
            \multirow{3}{*}{\makecell{Category-level\\retrieval Top-5}} 
            & Phase-I & 7.5 & 4.0 & 6.0 & 6.0 & 3.0 & 5.0 & 3.5 & 5.0 & 3.5 & 4.0 & 4.8 \\
            & Phase-II & 19.0 & 18.0 & 21.5 & 23.5 & 14.0 & 24.0 & 24.5 & 24.5 & 20.0 & 16.5 & 20.6 \\
            & Phase-III  & 28.5 & 35.5 & 40.5 & 43.5 & 34.5 & 43.0 & 40.5 & 45.5 & 36.0 & 46.5 & 39.4 \\
            
            \midrule
			\multirow{3}{*}{\makecell{Category-level\\retrieval Top-10}} 
            & Phase-I & 9.0 & 7.5 & 9.5 & 11.5 & 9.5 & 9.5 & 7.0 & 10.5 & 8.5 & 7.0 & 9.0 \\
            & Phase-II & 25.5 & 28.5 & 32.0 & 34.0 & 23.0 & 38.0 & 32.5 & 34.5 & 28.0 & 32.0 & 30.8 \\
            & Phase-III  & 39.0 & 49.5 & 51.0 & 57.0 & 48.5 & 55.5 & 54.5 & 57.5 & 49.0 & 61.5 & 52.3 \\
            \midrule
            \midrule
			\multirow{3}{*}{\makecell{Global\\retrieval Top-1}} 
            & Phase-I & 0.0 & 0.0 & 0.0 & 0.5 & 0.0 & 0.0 & 0.0 & 0.0 & 0.5 & 0.5 & 0.2 \\
            & Phase-II & 1.0 & 0.5 & 1.0 & 2.0 & 1.0 & 1.0 & 0.0 & 1.5 & 1.0 & 0.0 & 0.9 \\
            & Phase-III  & 3.0 & 4.5 & 8.0 & 5.0 & 4.5 & 8.0 & 5.5 & 9.0 & 5.0 & 6.5 & 5.9 \\
            \midrule
			\multirow{3}{*}{\makecell{Global\\retrieval Top-5}} 
            & Phase-I & 1.0 & 1.0 & 1.0 & 0.5 & 0.5 & 1.0 & 0.0 & 1.0 & 0.5 & 0.5 & 0.7 \\
            & Phase-II & 4.0 & 2.5 & 4.5 & 4.0 & 3.5 & 5.5 & 4.5 & 4.5 & 4.5 & 1.5 & 3.9 \\
            & Phase-III  & 15.0 & 16.5 & 18.5 & 17.0 & 17.0 & 24.5 & 20.5 & 28.5 & 15.0 & 23.5 & 19.6 \\
            \midrule
			\multirow{3}{*}{\makecell{Global\\retrieval Top-10}} 
            & Phase-I & 1.0 & 1.5 & 1.0 & 1.0 & 1.0 & 1.5 & 0.5 & 1.5 & 1.0 & 1.0 & 1.1 \\
            & Phase-II & 6.5 & 5.5 & 6.5 & 10.5 & 6.5 & 9.5 & 9.5 & 9.0 & 8.5 & 7.5 & 8.0 \\
            & Phase-III  & 19.0 & 26.0 & 28.0 & 29.0 & 24.0 & 34.5 & 30.0 & 38.0 & 23.0 & 34.5 & 28.6 \\
            \bottomrule
        \end{tabular}%
    }
\end{table*}
The quantitative results are reported in Table~\ref{tab:expanded-multi-image}. Several important observations can be drawn from this more challenging evaluation. First, although retrieval becomes substantially harder when the gallery is expanded from the standard one-image-per-class setting to a 2,400-image search space, the proposed framework still preserves meaningful retrieval capability across both category-level retrieval and global exact-image retrieval. This indicates that the learned EEG representations are not limited to coarse class discrimination under the simplified benchmark, but retain discriminative value in the presence of substantial within-class visual variation.

Second, the three representation levels exhibit a clear hierarchical progression. The \textbf{Phase-I} representation yields the weakest performance across all metrics, indicating that low-level perceptual representations alone are insufficient for resolving the increased ambiguity introduced by multiple instances per class. The \textbf{Phase-II} representation substantially improves both category-level retrieval and global exact-image retrieval, suggesting that high-level semantic learning already provides a more discriminative representation space than shallow perceptual alignment. The \textbf{Phase-III} representation further and consistently outperforms both \textbf{Phase-I} and \textbf{Phase-II} on all reported metrics, showing that the final fused representation is not merely a larger embedding, but a more effective integration of complementary perceptual and semantic information.

Third, the observed stage-wise gains provide stronger evidence for the rationale of the proposed framework than the standard benchmark alone. In particular, category-level retrieval mainly reflects whether the learned EEG representation can preserve class-discriminative structure under enlarged visual variability, whereas global exact-image retrieval additionally tests whether the representation can resolve fine-grained instance ambiguity within the target class. The consistent improvement from \textbf{Phase-I} to \textbf{Phase-II} and then to \textbf{Phase-III} across both evaluation perspectives suggests that the proposed framework progressively transforms EEG signals from shallow perceptual representations into more structured and discriminative semantic representations, and finally into a unified embedding that is better suited for fine-grained retrieval.

Overall, the expanded multi-image evaluation supports the central claim of this work that EEG visual decoding benefits from explicitly staged representation learning. Rather than relying on a single undifferentiated embedding space, the proposed framework forms a hierarchy of complementary EEG representations whose integration yields the strongest performance under a substantially more demanding retrieval setting.

\subsubsection{Qualitative Retrieval Refinement Analysis}
\begin{figure*}
    \centering
    \includegraphics[width=0.98\textwidth]{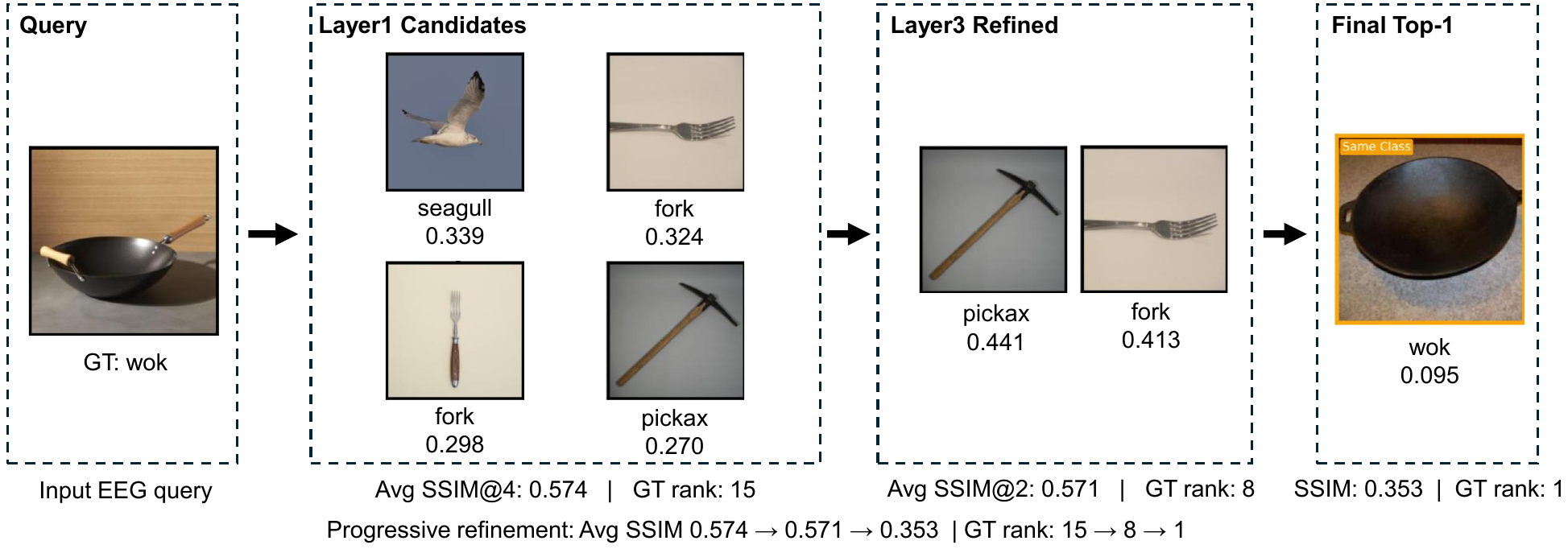}
    \caption{Qualitative retrieval refinement example on the expanded multi-image gallery. For the EEG query corresponding to the class \emph{wok}, the retrieval results show progressive refinement across the three stages. The Phase-I representation retrieves visually similar but semantically ambiguous candidates, the Phase-II representation narrows the results toward more category-relevant objects, and the Phase-III fused representation retrieves the target class at Top-1. The accompanying SSIM values, computed with respect to the query stimulus, suggest that earlier-stage retrieval is more influenced by low-level structural similarity, whereas the final fused representation appears less dominated by superficial visual resemblance and more guided by semantically discriminative information. For visualization clarity, more candidates are displayed at earlier stages and fewer at later stages (Phase-I: top-4, Phase-II: top-2, Phase-III: top-1), emphasizing the progressive concentration of retrieval results rather than providing a strictly matched top-$k$ quantitative comparison.}
    \label{fig:progressive_refinement}
\end{figure*}

To further understand the retrieval dynamics across representation levels, we provide a qualitative example in Fig.~\ref{fig:progressive_refinement}. For the EEG query corresponding to the class \emph{wok}, the retrieval results exhibit a clear progressive refinement across stages. The \textbf{Phase-I} representation is dominated by visually similar yet semantically ambiguous candidates, suggesting that it is primarily driven by low-level perceptual resemblance. The \textbf{Phase-II} representation narrows the candidate set toward more category-relevant objects, reflecting a stronger semantic concentration. Finally, the \textbf{Phase-III} representation ranks the correct class at Top-1, illustrating how the proposed staged framework progressively transforms shallow visual similarity into more discriminative category-level semantic identification.

The accompanying SSIM values further support this interpretation. Earlier-stage retrieval results tend to have relatively higher structural similarity to the query image, whereas the Phase-III retrieval places less emphasis on superficial visual resemblance and more on semantically discriminative information. For visualization clarity, we display more candidates at earlier stages and fewer at later stages (Phase-I: top-4, Phase-II: top-2, Phase-III: top-1), so as to emphasize the progressive concentration of retrieval results rather than to provide a strictly top-\(k\)-matched quantitative comparison.

Overall, this qualitative example provides intuitive support for the progressive refinement behavior of the proposed framework. Rather than producing a single undifferentiated embedding, the model forms a hierarchy of representations with complementary roles: early-stage features preserve perceptual similarity, high-level features enhance semantic discrimination, and the final fused representation integrates both to achieve the most accurate retrieval.

\subsection{Ablation Study}
\label{sec:ablation_study}

To validate the rationale behind each design of our framework, we conduct the following ablation experiments:

\begin{itemize}[itemsep=0pt, topsep=0pt]
    \item \textbf{Ours-All}: The full version of our proposed method with all components enabled. This setting serves as the reference model for evaluating the contribution of each individual design.

    \item \textbf{xC-Ori-12}: The proposed 12 semantic latent channels (i.e., $\mathbf{e}_{\mathrm{slc}}$ in Section~3.2.1) are removed and replaced with 12 real language-related semantic EEG channels. This ablation is designed to test whether the proposed learnable semantic latent channels provide a more effective semantic carrier than directly reusing real language-related channels.

    \item \textbf{xC-Ori-12-xC}: Based on \textbf{xC-Ori-12}, the coarse semantic branch is further disabled, i.e., Eq.~\ref{eq:phaseII-coarse} and Eq.~\ref{eq:clip_II_coarse} are removed. This setting is used to examine whether coarse semantic supervision still contributes when the proposed semantic latent channels are replaced by real language-related channels.

    \item \textbf{xC-xSLC}: The 12 semantic latent channels are removed, and only the 17 visual EEG channels are retained. In this case, the coarse semantic branch in Eq.~\ref{eq:phaseII-coarse} is computed using features derived from the retained visual channels. This ablation is used to evaluate the net benefit of semantic latent channel augmentation beyond using only the observed visual EEG channels.

    \item \textbf{xP-xPhaseI}: Phase-I is disabled. This setting is designed to assess the contribution of low-level visual representation learning to the overall staged framework.

    \item \textbf{xP-xPhaseII}: Phase-II is disabled. This ablation is used to evaluate the overall importance of structured semantic representation learning in the proposed framework.

    \item \textbf{xP-PhaseII-xF}: The fine semantic branch in Phase-II is disabled, i.e., Eq.~\ref{eq:phaseII_fine} and Eq.~\ref{eq:clip_II_fine} are removed. This setting is designed to test the contribution of fine image-level semantic modeling within Phase-II.

    \item \textbf{xP-PhaseII-xC}: The coarse semantic branch in Phase-II is disabled, i.e., Eq.~\ref{eq:phaseII-coarse} and Eq.~\ref{eq:clip_II_coarse} are removed. This ablation is used to assess the role of coarse label-level semantic supervision within Phase-II.
\end{itemize}

\subsubsection{Subject-Dependent Ablation}
\begin{table*}
    \centering
    \caption{Ablation studies on subject-dependent Top-1 and Top-5 accuracy (\%).}
    \label{tab:sub-dependent-ab}
    \renewcommand{\arraystretch}{1.1}
    \resizebox{\textwidth}{!}
    {%
    \begin{tabular}{llccccccccccc}
        \toprule
        \textbf{Metrics}& \textbf{Variants} & \textbf{Sub01} & \textbf{Sub02} & \textbf{Sub03} & \textbf{Sub04} & \textbf{Sub05} & \textbf{Sub06} & \textbf{Sub07} & \textbf{Sub08} & \textbf{Sub09} & \textbf{Sub10} & \textbf{Avg.}\\
        \midrule
        \multirow{8}{*}{\makecell{Top-1\\Accuracy}} 
        & Ours-All       &48.5            &\textbf{56.0}    &\textbf{53.5}    &\textbf{54.0}    & {44.0} &\textbf{60.0}    &\textbf{51.5}    &64.0            &\textbf{52.5}    & {66.0} &\textbf{55.0} \\
        & xC-Ori-12      &45.5            &51.5             &46.5             &51.0             &42.5             &55.5             &42.0             &61.5             &47.0             &59.5             &50.3 \\
        & xC-Ori-12-xC   &43.0            &48.0             &47.5             & 50.5 &41.5             &54.5             &42.0             &59.5             &43.5             &60.0             &49.0 \\
        & xC-xSLC        &45.0            &53.5             & 51.0 & 52.5 &\textbf{45.5}    &57.0             & 50.5 &\textbf{67.0}    & {51.0} &61.0             &53.4 \\
        & xP-xPhaseI     &45.5            &52.5             & 50.0 &50.0             &\textbf{45.5}    &56.0             &48.0             & 64.0 &49.0             &63.5             &52.4 \\
        & xP-xPhaseII    &46.0            &45.0             &50.5             &50.5             &42.0             &54.5             &47.0             &54.5             &42.5             &54.5             &48.7 \\
        & xP-PhaseII-xF  &47.0&44.0             &45.5             &47.0             &38.0             &55.0             &44.5             &57.5             &42.0             &57.0             &47.8 \\
        & xP-PhaseII-xC  &\textbf{51.0}   & {55.0} & {51.0} &49.5             &\textbf{45.5}    & 59.0 & 50.5 &63.5             &49.5             &\textbf{66.5}    & 54.1 \\
        \midrule
        \midrule
		\multirow{8}{*}{\makecell{Top-5\\Accuracy}} 
        & Ours-All       &74.0            &\textbf{87.5}    &\textbf{88.0}    &80.0             &\textbf{79.5}    &\textbf{88.0}    &\textbf{83.0}    & {89.0} & {81.5} &\textbf{91.0}    &\textbf{84.2}\\
        & xC-Ori-12      & {77.5}&81.0             &84.0             &\textbf{83.5}    &76.0             &81.5             &80.0             &89.5             &81.5             &89.5             &82.4\\
        & xC-Ori-12-xC   &73.5            &81.5             &81.0             & {83.0} &75.5             &82.0             &77.5             &86.0             &79.5             &86.0             &80.6\\
        & xC-xSLC        &75.5            &84.5             & {87.5} &80.5             &77.0             &85.0             &81.0             & {89.0} &\textbf{83.5}    &88.0             &83.2\\
        & xP-xPhaseI     &70.5            & 85.0 &84.5             &77.5             &76.5             & 84.5 &80.5             & {89.0} &\textbf{83.5}    &86.5             &81.8\\
        & xP-xPhaseII    &\textbf{78.0}   &80.5             &79.0             &77.0             &72.5             &81.5             &77.0             &83.0             &78.0             &86.0             &79.3\\
        & xP-PhaseII-xF  & {76.0}&81.5             &82.5             &82.5             &71.5             &83.5             &76.5             &85.0             &79.0             &86.0             &80.4\\
        & xP-PhaseII-xC  &73.5            &84.5             &87.0             &80.5             & {78.5} &86.0             & {82.5} &\textbf{90.0}    &80.0             & {90.5} & {83.3}\\
        \bottomrule
    \end{tabular}%
    }
\end{table*}

We conduct ablation studies in the subject-dependent 200-way zero-shot setting. Table~\ref{tab:sub-dependent-ab} summarises the contribution of each component.
The full model (\textbf{Ours-All}) attains the highest Top-1/Top-5 averages of \textbf{55.0\%}/\textbf{84.2\%}. 
Replacing the proposed \emph{semantic latent channels} with the real 12 language-related semantic channels (\textbf{xC-Ori-12}) reduces Top-1 to \textbf{50.3\%} (-4.7) and Top-5 to \textbf{82.4\%} (-1.8), indicating that the semantic latent channels provide a stronger semantic carrier for visual EEG.
Removing the coarse semantic branch on top of this replacement (\textbf{xC-Ori-12-xC}) further drops performance to \textbf{49.0\%}/\textbf{80.6\%} (Top-1/Top-5), showing that coarse label-level supervision contributes to robustness beyond fine-grained cues.
When the semantic latent channels are removed but only the 17 visual channels are kept (\textbf{xC-xSLC}), the model performs reasonably (\textbf{53.4\%}/\textbf{83.2\%}), yet lags behind \textbf{Ours-All} by 1.6/1.0 points (Top-1/Top-5), confirming the net gain brought by semantic latent channel augmentation.

Eliminating the low-level phase (\textbf{xP-xPhaseI}) leads to \textbf{52.4\%}/\textbf{81.8\%}, evidencing the necessity of early-stage low-level alignment for stabilizing mid-/late-stage learning.
Removing the dual-level semantic phase altogether (\textbf{xP-xPhaseII}) causes a larger degradation to \textbf{48.7\%}/\textbf{79.3\%}, underscoring the central role of Phase-II.
Within Phase-II, disabling the fine semantic branch (\textbf{xP-PhaseII-xF}) yields \textbf{47.8\%}/\textbf{80.4\%}, whereas disabling the coarse semantic branch (\textbf{xP-PhaseII-xC}) gives \textbf{53.0\%}/\textbf{83.3\%}.
Thus, the \emph{fine semantic} branch is the primary driver for Top-1 discrimination (largest drop when removed), while the \emph{coarse semantic} branch improves calibration/recall (clear Top-5 gain), and their combination with Phase-I features (Phase-III fusion) delivers the best overall accuracy and consistency.

Overall, these results support the rationale of the proposed staged design. Phase-I helps anchor low-level visual information, Phase-II contributes structured semantic modeling through dual-level supervision and semantic latent channels, and Phase-III integrates these complementary representations into a unified EEG embedding. Their combination achieves the best average performance among the ablation variants in the subject-dependent setting, confirming the effectiveness of the proposed framework and its semantic modeling strategy.

\subsubsection{Subject-Independent Ablation}

\begin{table*}
    \centering
    \caption{Ablation studies on subject-independent Top-1 and Top-5 accuracy (\%) in 200-way zero-shot.}
    \label{tab:appendix-sub-independent-ab}
    \renewcommand{\arraystretch}{1.1}
    \resizebox{\textwidth}{!}
    {%
    \begin{tabular}{llccccccccccc}
        \toprule
        \textbf{Metrics}& \textbf{Variants} & \textbf{Sub01} & \textbf{Sub02} & \textbf{Sub03} & \textbf{Sub04} & \textbf{Sub05} & \textbf{Sub06} & \textbf{Sub07} & \textbf{Sub08} & \textbf{Sub09} & \textbf{Sub10} & \textbf{Avg.}\\
        \midrule
		\multirow{8}{*}{\makecell{Top-1\\Accuracy}} 
        & Ours-All           & 13.0 & \textbf{16.5} & \textbf{8.0} & \textbf{14.5} & 10.0 & 14.0 & 9.5 & \textbf{11.5}& \textbf{14.5} & \textbf{20.5} & \textbf{13.2} \\
        & xC-Ori-12      & 11.5 & 16.0 		  & 7.0 		 & 14.0 		 & 10.0 & 12.5 & 9.0 & 11.0 		& 13.5 			& 17.5 & 12.2\\
        & xC-Ori-12-xC   & 11.0 & 16.0         & 6.0 		 & 12.5 		 & 9.5  & 11.5 & 8.0 & 10.5 		& 12.0 			& 15.5 &  11.3\\
        & xC-xSLC        & 14.0 & 16.0 & 5.5 & 14.0 & \textbf{13.0} & 11.5 & \textbf{10.0} & \textbf{11.5} & 11.5 & 15.0 & 12.3 \\
        & xP-xPhaseI     & \textbf{15.0} & 14.5 & 4.0 & 14.0 & 12.0 & \textbf{14.5} & 2.0  & 10.0 & 13.5 & 17.5 & 11.7 \\
        & xP-xPhaseII    & 10.5 & 12.5 & 5.5 & 14.0 & 8.0  & 13.0 & 7.5  & 10.0 & 13.0 & 15.0 & 10.9 \\
        & xP-PhaseII-xF  & 12.0 & 14.0 & 7.5 & 12.0 & 7.0  & 11.5 & 7.0  & 8.5  & 10.5 & 14.5 & 10.5 \\
        & xP-PhaseII-xC  & 13.0 & 15.5 & 5.5 & \textbf{14.5} & 9.0  & 11.0 & 7.5  & 8.0  & 9.5  & 16.0 & 11.0 \\
        \midrule
        \midrule
		\multirow{8}{*}{\makecell{Top-5\\Accuracy}}
        & Ours-All       & 32.0 & 41.5 & 22.0 & 34.5 & 31.5 & 31.5 & \textbf{27.0} & 30.5 & \textbf{32.0} & 40.0 & \textbf{32.3} \\
        & xC-Ori-12      & 34.0 & \textbf{43.0} & 22.0 & 32.0 & 26.0 & 33.0 & 23.0 & \textbf{33.0} & 31.5 & 42.0 & 32.0 \\
        & xC-Ori-12-xC   & 33.5 & 41.5 & \textbf{23.0} & 32.0 & 30.5 & 32.5 & 22.0 & 30.5 & 31.0 & 42.5 & 31.9 \\
        & xC-xSLC        & 31.5 & 40.0 & 20.0 & 31.0 & \textbf{32.0} & 33.5 & 24.0 & 32.0 & 30.5 & 40.5 & 31.5 \\
        & xP-xPhaseI     & \textbf{34.5} & 38.5 & 15.0 & \textbf{35.0} & 31.5 & \textbf{35.5} & 18.5 & 29.5 & 29.0 & \textbf{44.5} & 31.2 \\
        & xP-xPhaseII    & 29.5 & 36.5 & 18.5 & 31.0 & 23.0 & 33.0 & 21.0 & 27.5 & 30.0 & 41.5 & 29.2 \\
        & xP-PhaseII-xF  & 27.5 & 41.5 & 18.5 & 30.0 & 24.5 & 28.5 & 17.5 & 26.5 & 26.0 & 41.0 & 28.2 \\
        & xP-PhaseII-xC  & 32.5 & 40.0 & 17.0 & 32.5 & 28.0 & 30.0 & 24.5 & \textbf{33.0} & 27.5 & 41.0 & 30.6 \\
        \bottomrule
    \end{tabular}%
    }
\end{table*}

In the more challenging inter-subject setting, the full model still achieves the strongest average performance in Table~\ref{tab:appendix-sub-independent-ab}, reaching 13.2\%/32.3\% in Top-1/Top-5. Overall, these results suggest that the proposed framework remains effective under subject shift, although the contribution of each component becomes more variable across subjects than in the subject-dependent setting.

Replacing the proposed semantic latent channels with real language-related semantic channels (xC-Ori-12) reduces the average Top-1 accuracy to 12.2\%, while Top-5 remains close to the full model at 32.0\%. This suggests that the learned semantic latent channels provide a more effective semantic carrier for exact cross-subject discrimination. When the coarse semantic branch is further removed on top of this replacement (xC-Ori-12-xC), the average performance drops further to 11.3\%/31.9\% in Top-1/Top-5, indicating that coarse semantic supervision still provides useful complementary support when combined with a weakened semantic carrier. When the semantic latent channels are entirely removed and only the visual channels are retained (xC-xSLC), performance remains competitive at 12.3\%/31.5\%, but still lags behind the full model, supporting the overall benefit of semantic latent channel augmentation.

Phase-level ablations further show that semantic modeling remains particularly important in the inter-subject scenario. Disabling Phase-I (xP-xPhaseI) yields 11.7\%/31.2\%, whereas removing Phase-II altogether (xP-xPhaseII) causes a larger degradation to 10.9\%/29.2\%, suggesting that the semantic stage contributes more substantially to cross-subject generalization. Within Phase-II, removing the fine semantic branch (xP-PhaseII-xF) leads to the weakest overall performance at 10.5\%/28.2\%, indicating that fine semantic discrimination plays a major role in exact retrieval under subject shift. In contrast, removing the coarse semantic branch (xP-PhaseII-xC) gives 11.0\%/30.6\%, which is clearly better than removing the fine branch but still below the full model, suggesting that the coarse branch mainly provides complementary support for broader semantic recall and retrieval stability.

Overall, the subject-independent ablation results suggest that cross-subject generalization benefits most from the joint effect of semantic latent channels, fine semantic refinement, and multi-stage integration, while the coarse semantic branch provides complementary support rather than being the primary driver of exact discrimination.

\subsection{Model Complexity Analysis}

\begin{table}
    \centering
    \caption{Comparison of model complexity.}
    \label{tab:model_complexity}
    \resizebox{\columnwidth}{!}
    {%
    \begin{tabular}{lccc}
        \toprule
        \textbf{Methods} & \textbf{Parameters (M)} & \textbf{MMACs} & \textbf{MFLOPs} \\
        \midrule
        \citet{du2023decoding}                    & 2.7  & 1.4  & 2.7   \\
        \citet{song2024decoding}                  & 2.5  & 4.4  & 9.1   \\
        \citet{li2024visual}                      & 2.5  & 9.4  & 19.3  \\
        \citet{zhang2025category}                 & 3.7  & 20.1 & 40.9  \\
        \citet{zhang2025cognitioncapturer}        & 13.5 & 54.0 & 108.8 \\
        \citet{wu2025bridging}                    & 4.1  & 4.1  & 8.2   \\
        Ours                                      & 10.6  & 12.2  & 24.5  \\
        \bottomrule
    \end{tabular}%
    }
\end{table}
As shown in Table~\ref{tab:model_complexity}, the proposed model contains 10.6M parameters, with 12.2 MMACs and 24.5 MFLOPs. Although it is not the smallest architecture among the compared methods, its computational cost remains moderate and substantially lower than heavier models such as \citet{zhang2025cognitioncapturer}. More importantly, the proposed method achieves superior retrieval performance without incurring the highest computational burden, suggesting a favorable trade-off between model complexity and effectiveness.

\vspace{-0.2cm}
\section{Conclusion}
\label{sec:con}
This work presents a neuroscience-inspired staged framework for EEG-based visual decoding by organizing EEG representation learning into low-level visual representation learning, high-level semantic representation learning, and integrative information fusion. Rather than treating EEG decoding as a single global alignment problem, the proposed framework shows that explicitly staged representation learning can improve within-subject decoding performance and strengthen exact cross-subject retrieval. By introducing dual-level multimodal semantic learning and semantic latent channels, the proposed method enhances the semantic representational capacity of EEG signals and supports more structured cross-modal alignment. Experiments and ablation studies on the THINGS-EEG benchmark confirm the effectiveness of the proposed design and highlight the complementary roles of low-level perception, structured semantic modeling, and integrative fusion. Overall, this study suggests that incorporating staged processing insights into EEG representation learning is a promising direction for improving EEG visual decoding. Future work may further strengthen cross-subject robustness and investigate closer links between computational stages and interpretable neural dynamics.

\printcredits

\section*{Data Availability}
The dataset used in this study, THINGS-EEG, is publicly available online. It can be accessed at \url{https://osf.io/3jk45/overview}.

\section*{Declaration of Competing Interest}
The authors declare that they have no known competing financial interests or personal relationships that could have appeared to influence the work reported in this paper.


\bibliographystyle{cas-model2-names}
\bibliography{neurocomputing}

@inproceedings{wu2025bridging,
  title={Bridging the Vision-Brain Gap with an Uncertainty-Aware Blur Prior},
  author={Wu, Haitao and Li, Qing and Zhang, Changqing and He, Zhen and Ying, Xiaomin},
  booktitle={CVPR},
  pages={2246--2257},
  year={2025}
}

@inproceedings{zhang2025cognitioncapturer,
  title={CognitionCapturer: Decoding Visual Stimuli From Human EEG Signal With Multimodal Information},
  author={Zhang, Kaifan and He, Lihuo and Jiang, Xin and Lu, Wen and Wang, Di and Gao, Xinbo},
  booktitle={AAAI},
  volume={39},
  number={13},
  pages={14486--14493},
  year={2025}
}

@inproceedings{zhang2025category,
  title={Category-aware EEG image generation based on wavelet transform and contrast semantic loss},
  author={Zhang, Enshang and others},
  booktitle={IJCAI},
  year={2025}
}

@article{li2024visual,
  title={Visual Decoding and Reconstruction via EEG Embeddings with Guided Diffusion},
  author={Li, Dongyang and Wei, Chen and Li, Shiying and Zou, Jiachen and Liu, Quanying},
  journal={NeurIPS},
  volume={37},
  pages={102822--102864},
  year={2024}
}

@inproceedings{song2024decoding,
  title={Decoding Natural Images from EEG for Object Recognition},
  author={Song, Yonghao and Liu, Bingchuan and Li, Xiang and Shi, Nanlin and Wang, Yijun and Gao, Xiaorong},
  booktitle={ICLR},
  year={2024}
}

@article{chen2024visual,
  title={Visual neural decoding via improved visual-EEG semantic consistency},
  author={Chen, Hongzhou and He, Lianghua and Liu, Yihang and Yang, Longzhen},
  journal={arXiv preprint arXiv:2408.06788},
  year={2024}
}

@article{du2023decoding,
  title={Decoding visual neural representations by multimodal learning of brain-visual-linguistic features},
  author={Du, Changde and Fu, Kaicheng and Li, Jinpeng and He, Huiguang},
  journal={IEEE TPAMI},
  volume={45},
  number={9},
  pages={10760--10777},
  year={2023},
  publisher={IEEE}
}

@inproceedings{ferrante2024decoding,
  title={Decoding eeg signals of visual brain representations with a clip based knowledge distillation},
  author={Ferrante, Matteo and Boccato, Tommaso and Bargione, Stefano and Toschi, Nicola},
  booktitle={ICLR 2024 Workshop on Learning from Time Series For Health},
  year={2024}
}

@inproceedings{ma2024cognition,
  title={Cognition-Supervised Saliency Detection: Contrasting EEG Signals and Visual Stimuli},
  author={Ma, Jun and Ruotsalo, Tuukka},
  booktitle={ACM MM},
  pages={7744--7753},
  year={2024}
}

@inproceedings{xiao2025eeg,
  title={EEG Decoding and Visual Reconstruction via 3D Geometric with Nonstationarity Modelling},
  author={Xiao, Xin and Wei, Kaiwen and Zhong, Jiang and Wei, Xuekai and Yan, Jielu},
  booktitle={ICASSP},
  pages={1--5},
  year={2025},
  organization={IEEE}
}

@inproceedings{rajabi2025human,
  title={Human-Aligned Image Models Improve Visual Decoding from the Brain},
  author={Rajabi, Nona and Ribeiro, Ant{\^o}nio H and Vasco, Miguel and Taleb, Farzaneh and Bj{\"o}rkman, M{\aa}rten and Kragic, Danica},
  booktitle={ICML},
  year={2025}
}

@article{song2025recognizing,
  title={Recognizing Natural Images From EEG With Language-Guided Contrastive Learning},
  author={Song, Yonghao and Wang, Yijun and He, Huiguang and Gao, Xiaorong},
  journal={IEEE Transactions on Neural Networks and Learning Systems},
  year={2025},
  publisher={IEEE}
}

@article{zeng2023dm,
  title={DM-RE2I: A framework based on diffusion model for the reconstruction from EEG to image},
  author={Zeng, Hong and Xia, Nianzhang and Qian, Dongguan and Hattori, Motonobu and Wang, Chu and Kong, Wanzeng},
  journal={Biomedical Signal Processing and Control},
  volume={86},
  pages={105125},
  year={2023},
  publisher={Elsevier}
}

@article{gifford2022large,
  title={A large and rich EEG dataset for modeling human visual object recognition},
  author={Gifford, Alessandro T and Dwivedi, Kshitij and Roig, Gemma and Cichy, Radoslaw M},
  journal={NeuroImage},
  volume={264},
  pages={119754},
  year={2022},
  publisher={Elsevier}
}

@article{kroczek2019contributions,
  title={Contributions of left frontal and temporal cortex to sentence comprehension: Evidence from simultaneous TMS-EEG},
  author={Kroczek, Leon OH and Gunter, Thomas C and Rysop, Anna U and Friederici, Angela D and Hartwigsen, Gesa},
  journal={Cortex},
  volume={115},
  pages={86--98},
  year={2019},
  publisher={Elsevier}
}

@article{felleman1991distributed,
  title={Distributed hierarchical processing in the primate cerebral cortex.},
  author={Felleman, Daniel J and Van Essen, David C},
  journal={Cerebral Cortex (New York, NY: 1991)},
  volume={1},
  number={1},
  pages={1--47},
  year={1991}
}

@article{kappenman2021erp,
  title={ERP CORE: An open resource for human event-related potential research},
  author={Kappenman, Emily S and Farrens, Jaclyn L and Zhang, Wendy and Stewart, Andrew X and Luck, Steven J},
  journal={NeuroImage},
  volume={225},
  pages={117465},
  year={2021},
  publisher={Elsevier}
}

@article{xu2021review,
  title={Review of brain encoding and decoding mechanisms for EEG-based brain--computer interface},
  author={Xu, Lichao and Xu, Minpeng and Jung, Tzyy-Ping and Ming, Dong},
  journal={Cognitive Neurodynamics},
  volume={15},
  number={4},
  pages={569--584},
  year={2021},
  publisher={Springer}
}

@article{goodale1992separate,
  title={Separate visual pathways for perception and action},
  author={Goodale, Melvyn A and Milner, A David},
  journal={Trends in Neurosciences},
  volume={15},
  number={1},
  pages={20--25},
  year={1992},
  publisher={Elsevier}
}

@article{graumann2022spatiotemporal,
  title={The spatiotemporal neural dynamics of object location representations in the human brain},
  author={Graumann, Monika and Ciuffi, Caterina and Dwivedi, Kshitij and Roig, Gemma and Cichy, Radoslaw M},
  journal={Nature Human Behaviour},
  volume={6},
  number={6},
  pages={796--811},
  year={2022},
  publisher={Nature Publishing Group UK London}
}

@software{ilharco_gabriel_2021_5143773,
  author       = {Ilharco, Gabriel and
                  Wortsman, Mitchell and
                  Wightman, Ross and
                  Gordon, Cade and
                  Carlini, Nicholas and
                  Taori, Rohan and
                  Dave, Achal and
                  Shankar, Vaishaal and
                  Namkoong, Hongseok and
                  Miller, John and
                  Hajishirzi, Hannaneh and
                  Farhadi, Ali and
                  Schmidt, Ludwig},
  title        = {OpenCLIP},
  month        = jul,
  year         = 2021,
  publisher    = {Zenodo},
  version      = {0.1},
  doi          = {10.5281/zenodo.5143773},
}

@article{pereira2018toward,
  title={Toward a universal decoder of linguistic meaning from brain activation},
  author={Pereira, Francisco and Lou, Bin and Pritchett, Brianna and Ritter, Samuel and Gershman, Samuel J and Kanwisher, Nancy and Botvinick, Matthew and Fedorenko, Evelina},
  journal={Nature Communications},
  volume={9},
  number={1},
  pages={963},
  year={2018},
  publisher={Nature Publishing Group UK London}
}

@article{ding2025eeg,
  title={EEG-based brain-computer interface enables real-time robotic hand control at individual finger level},
  author={Ding, Yidan and Udompanyawit, Chalisa and Zhang, Yisha and He, Bin},
  journal={Nature Communications},
  volume={16},
  number={1},
  pages={1--20},
  year={2025},
  publisher={Nature Publishing Group}
}

@article{wilson2024feasibility,
  title={Feasibility of decoding visual information from EEG},
  author={Wilson, Holly and Chen, Xi and Golbabaee, Mohammad and Proulx, Michael J and O’Neill, Eamonn},
  journal={Brain-Computer Interfaces},
  volume={11},
  number={1-2},
  pages={33--60},
  year={2024},
  publisher={Taylor \& Francis}
}

@article{ferrante2024decoding_vs,
  title={Decoding visual brain representations from electroencephalography through knowledge distillation and latent diffusion models},
  author={Ferrante, Matteo and Boccato, Tommaso and Bargione, Stefano and Toschi, Nicola},
  journal={Computers in Biology and Medicine},
  volume={178},
  pages={108701},
  year={2024},
  publisher={Elsevier}
}

@inproceedings{zhang2025cat,
  title={CAT-Net: A Co-Adaptive Transfer Learning Network for BCI-Assisted Neurorehabilitation},
  author={Zhang, Shuailei and Ding, Yi and Jiang, Muyun and Tang, Ning and Chew, Effie and Ang, Kai Keng and Guan, Cuntai},
  booktitle={ICASSP},
  pages={1--5},
  year={2025},
  organization={IEEE}
}

@inproceedings{de2024perceptual,
  title={Perceptual visual similarity from EEG: Prediction and image generation},
  author={De La Torre-Ortiz, Carlos and Ruotsalo, Tuukka},
  booktitle={ACM MM},
  pages={11146--11155},
  year={2024}
}

@inproceedings{shi2025brainalign,
  title={BrainAlign: EEG-Vision Alignment via Frequency-Aware Temporal Encoder and Differentiable Cluster Assigner},
  author={Shi, E. and others},
  booktitle={MICCAI},
  pages={98--108},
  year={2025},
  organization={Springer}
}

@article{thigpen2017assessing,
  title={Assessing the internal consistency of the event-related potential: An example analysis},
  author={Thigpen, Nina N and Kappenman, Emily S and Keil, Andreas},
  journal={Psychophysiology},
  volume={54},
  number={1},
  pages={123--138},
  year={2017},
  publisher={Wiley Online Library}
}

@article{creel2019visually,
  title={Visually evoked potentials},
  author={Creel, Donnell Joseph},
  journal={Handbook of clinical neurology},
  volume={160},
  pages={501--522},
  year={2019},
  publisher={Elsevier}
}

@article{jing2025pinpointing,
  title={Pinpointing visual content: Disentangled features in multimodal model for eeg representation learning and decoding},
  author={Jing, Haodong and Ma, Yongqiang and Yang, Panqi and Hua, Haibo and Zheng, Nanning},
  journal={Knowledge-Based Systems},
  pages={114212},
  year={2025},
  publisher={Elsevier}
}

\end{document}